\documentclass[11pt]{article}

\usepackage[preprint]{acl}

\usepackage{acl}
\usepackage{times}
\usepackage{latexsym}

\usepackage{xcolor}
\usepackage{mdframed}
\usepackage{listings}

\usepackage{booktabs}
\usepackage{booktabs}
\usepackage{multirow}
\usepackage{makecell}

\usepackage{url}
\usepackage{graphicx}
\usepackage{amsmath} 
\usepackage{multirow}
\usepackage{authblk}
\usepackage[table]{xcolor} 
\usepackage{tcolorbox} 
\usepackage{enumitem}    
\usepackage{textcomp}   
\usepackage{hyperref}   
\hypersetup{
	colorlinks=true,
	linkcolor=red,   
	urlcolor=blue,
}

\usepackage{listings}   
\usepackage{xcolor}
\usepackage{makecell}

\definecolor{backblue}{RGB}{210, 230, 250}
\definecolor{backgreen}{RGB}{226, 240, 217}
\newcommand{\highblue}{\cellcolor{backblue}}
\newcommand{\highgreen}{\cellcolor{backgreen}}

\usepackage{xcolor}
\usepackage{pifont}
\newcommand{\greencheck}{\textcolor{green!70!black}{\ding{51}}}
\newcommand{\redxmark}{\textcolor{red}{\ding{55}}}

\usepackage[T1]{fontenc}

\usepackage[utf8]{inputenc}

\usepackage{microtype}

\usepackage{inconsolata}

\usepackage{graphicx}

\title{MemGround: Long-Term Memory Evaluation Kit for Large Language Models in Gamified Scenarios}

\author{
 \textbf{Yihang Ding\textsuperscript{1}\thanks{Equal Contribution.}},
 \textbf{Wanke Xia\textsuperscript{1}\footnote[1]{}},
 \textbf{Yiting Zhao\textsuperscript{1}\footnote[1]{}},
 \textbf{Jinbo Su\textsuperscript{2}},
\\
 \textbf{Jialiang Yang\textsuperscript{1}},
 \textbf{Zhengbo Zhang\textsuperscript{3}},
 \textbf{Ke Wang\textsuperscript{2}},
 \textbf{Wenming Yang\textsuperscript{1}\thanks{Corresponding Author.}}
\\
\\
 \textsuperscript{1}Tsinghua University,
 \textsuperscript{2}Renmin University of China,
 \textsuperscript{3}CASIA
\\
 \texttt{\{ding-yh25,xwk25,yiting-z25\}@mails.tsinghua.edu.cn}
\\
 \texttt{yang.wenming@sz.tsinghua.edu.cn}
}

\begin{document}
\maketitle
\begin{abstract}
Current evaluations of long-term memory in LLMs are fundamentally static. By fixating on simple retrieval and short-context inference, they neglect the multifaceted nature of complex memory systems, such as dynamic state tracking and hierarchical reasoning in continuous interactions. To overcome these limitations, we propose MemGround, a rigorous long-term memory benchmark natively grounded in rich, gamified interactive scenarios. To systematically assess these capabilities, MemGround introduces a three-tier hierarchical framework that evaluates Surface State Memory, Temporal Associative Memory, and Reasoning-Based Memory through specialized interactive tasks. Furthermore, to comprehensively quantify both memory utilization and behavioral trajectories, we propose a multi-dimensional metric suite comprising Question-Answer Score (QA Overall), Memory Fragments Unlocked (MFU), Memory Fragments with Correct Order (MFCO), and Exploration Trajectory Diagrams (ETD). Extensive experiments reveal that state-of-the-art LLMs and memory agents still struggle with sustained dynamic tracking, temporal event association, and complex reasoning derived from long-term accumulated evidence in interactive environments.
\footnote{Code and data will be available at \url{https://github.com/MorleyOlsen/MemGround}.}
\end{abstract}

\section{Introduction}
\begin{figure}[t]
    \centering
    \includegraphics[width=1\linewidth]{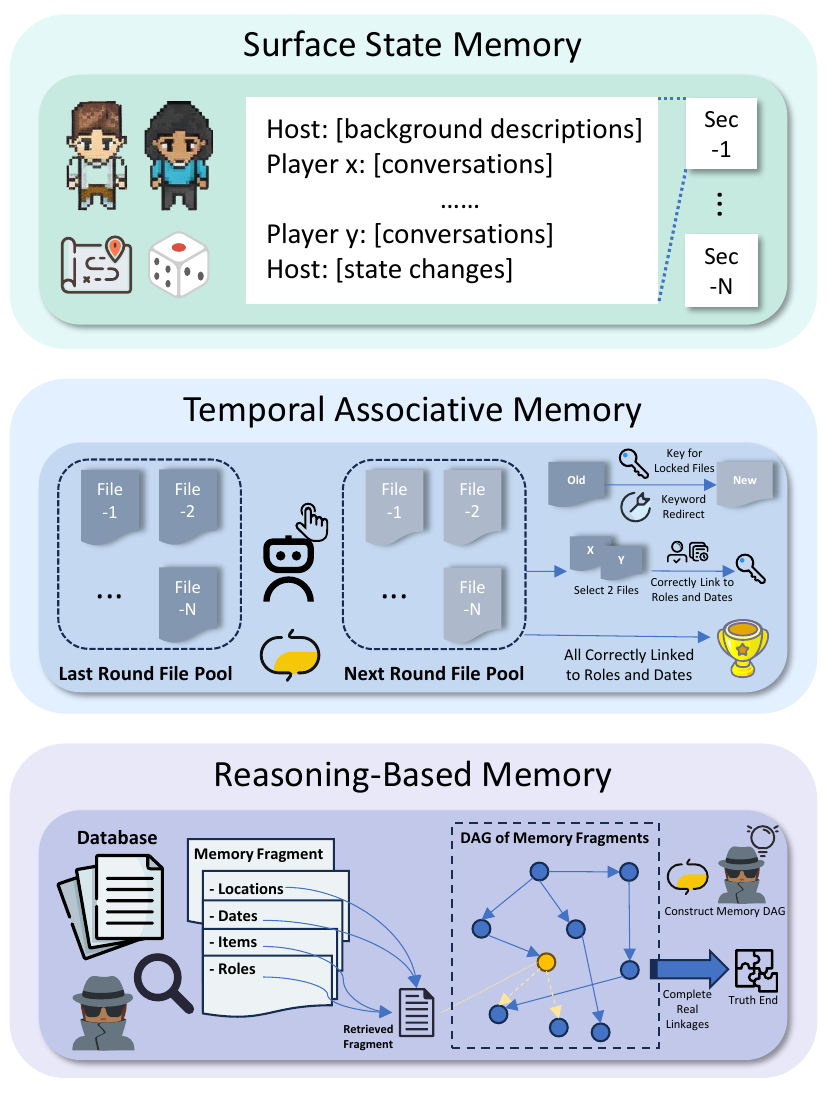}
    \caption{\textbf{Overview of MemGround.} MemGround includes three-tier hierachical memory evaluation framework, consisting of Surface State Memory, Temporal Associative Memory, and Reasoning-Based Memory. Each is evaluated in a particular gamified scenario.}
    \label{fig:1}
\end{figure}

The rapid advancement of Large Language Models (LLMs) and agents increasingly relies on their ability to manage long-term memory and execute complex reasoning tasks~\citep{zhong2024memorybank,liu2023think,hatalis2023memory,chang2025grail,wang2025asma}. 
As these models are widely deployed in interactive applications, such as intelligent assistants~\citep{dong2023towards}, virtual characters~\citep{wang2025characterbox}, and complex problem-solving environments~\citep{ma2025problem}, the demands on their memory capabilities have evolved significantly. 
Therefore, the evaluation of long-term memory for these systems is no longer limited to "\textbf{whether information is remembered}", but more focused on "\textbf{whether memory and reasoning are dynamic, hierarchical, and associative}".

However, existing benchmarks for LLMs' long-term memory and reasoning tend to emphasize static retrieval or short-context multi-hop QA~\citep{yang-etal-2018-hotpotqa,hsieh2024ruler,wu2025longmemeval,maharana2024lococmo}. 
While these datasets and tasks are valuable for measuring factual coverage and local chaining, they leave several critical gaps.
First, many evaluations assume that \textbf{all relevant information is present and explicitly accessible at query time}, which understates the difficulty of discovering and assembling information that is distributed across an evolving interaction. 
Second, most tasks \textbf{reduce evaluation to outcome correctness} (e.g., answer matches), offering limited visibility into the intermediate processes of memory retrieval, dependency resolution, and exploration strategy. 
Third, the majority of benchmarks are \textbf{linear or conversation-like in structure} and do not capture the graph-structured, branching dependencies that characterize real-world investigative and multi-agent scenarios. 
As a result, current evaluation suites provide only a partial picture of a model’s capacity to manage the kinds of dynamic, hierarchical, and associative memory operations.


Motivated by these limitations, we introduce \textbf{MemGround}, a long-term memory evaluation kit built around gamified interactive scenarios. 
Games provide a compact, well-specified substrate in which memory nodes, dependencies, and progression mechanics are explicit yet richly structured~\citep{carstensdottir2019player}.
Technically, MemGround is comprised of two tightly coupled components: (1) a human-grounded data construction pipeline that systematically extracts logic and dependency traces from real gameplay into structured JSON formats, and (2) a unified interactive evaluation framework that simulates dynamic agent-environment loops. 
To provide a fine-grained assessment, we further propose a three-tier hierarchical taxonomy targeting Surface State Memory, Temporal Associative Memory, and Reasoning-Based Memory. 
This approach shifts the evaluation paradigm from static retrieval to a dynamic process, stressing both the discovery process (how models explore and expand accessible memory) and the reasoning process (how models synthesize fragmented evidence into coherent conclusions).
We evaluate multiple closed-source and open source models and LLM-based memory agents to learn the long-term memory and reasoning capacities of LLMs and their agents.
Experiments show that while current models can often retrieve or follow instructions from previously observed information, they still struggle to consistently maintain coherent long-term memory and to perform reliable reasoning over distributed memory fragments across extended interactions. 

In conclusion, our main contributions can be summarized as follows:
\begin{itemize}
    \item We introduce MemGround, a novel long-term memory evaluation benchmark for LLMs with a three-tier hierarchical memory evaluation framework in gamified scenarios, including Surface State Memory, Temporal Associative Memory, and Reasoning-Based Memory.
    \item We further propose multi-dimensional evaluation metrics tailored to the three-tier hierarchical memory, which are Question-Answer Score, Memory Fragments Unlocked, Memory Fragments with Correct Order, and Exploration Trajectory Diagram of Subsections.
    \item We conduct detailed experiments of representative  closed-source and open-source models and LLM-based memory agents, verifying that state-of-the-art models still struggle with sustained tracking of dynamic memory, and logical reasoning based on long-term memory.
\end{itemize}

\section{Related Works}
\subsection{Memory Capabilities of LLMs and Agents}
Memory is the core capability enabling LLMs and agents to achieve agentic intelligence and support complex reasoning and interactive decision-making~\citep{wu2025human,hu2026memoryageaiagents}. 
Optimizing memory capabilities of LLMs and agents primarily falls into 2 categories: (1) Static memory enhancement, which suggests improving LLMs' retrieval and recall of static information through mechanisms like RAG and memory caching~\citep{yu2025can,gutierrez2025from,qian2025memorag,jimenez2024hipporag,su2025racqc}; (2) Dynamic memory exploration, which refers to design dynamic memory update mechanisms, enabling LLMs to track information changes in various scenarios~\citep{shinn2023reflexion,xu2025mem,wang2025mem,chhikara2025mem0}.


\subsection{Long-Term Memory Evaluation Benchmarks}
As research on memory capabilities of LLMs and agents advances, various benchmarks for long-term memory evaluation have been proposed. 
HotpotQA~\citep{yang-etal-2018-hotpotqa} introduces a Wikipedia-based multi-hop question answering dataset that has become a widely adopted benchmark for evaluating multi-step factual reasoning. 
RULER~\citep{hsieh2024ruler} further expands the evaluation scope by including multiple tasks such as retrieval and multi-hop tracking, enabling systematic analysis of long-context processing abilities. 
LongMemEval~\citep{wu2025longmemeval} focuses on conversational settings and proposes a benchmark covering several memory-related capabilities, including information extraction and multi-turn reasoning under scalable dialogue histories. 
Similarly, LoCoMo~\citep{maharana2024lococmo} constructs a multimodal long-dialogue dataset and evaluates models across tasks such as question answering, event summarization, and multimodal dialogue generation.
Though existing benchmarks have advanced long-term memory evaluation, they mainly focus on static multi-hop reasoning or long conversations, emphasizing retrieval and reasoning within fixed contexts. 
However, modern agents operate in interactive environments where states evolve and information is progressively discovered, making structured and dynamic evaluations increasingly necessary. 

\section{MemGround Construction}

\subsection{Environment Overview}

MemGround's design logic centers on providing an interactive, traceable, and dynamically evolving evaluation platform for three-tier memory capabilities.
We then apply \textit{TRPG}, \textit{No Case Should Remain Unsolved}, and \textit{Type Help} scenarios to evaluate Surface State Memory, Temporal Associative Memory, and Reasoning-Based Memory, respectively.

\subsubsection{TRPG}
\textit{TRPG} is a tabletop role-playing game that relies on verbal descriptions and typically requires character sheets and dice.
Players are mainly divided into two roles: hosts and performers. Hosts act as narrators and arbiters who describe the world, characters, and encounters, while performers play specific roles and interact with hosts and other performers. The outcomes of encounters depend on the performers’ attributes and actions, the hosts’ world design, and objective factors represented by dice rolls.

This game primarily evaluates \textbf{Surface State Memory}, which refers to the ability to continuously track observable states during interaction. 
Abstractly, the interaction follows a state tracking paradigm in which the environment presents a sequence of observable states that evolve as the interaction progresses. 
The model must therefore maintain an up-to-date representation of the current environment state by remembering previously introduced variables and updating them when new events occur.
This paradigm naturally arises due to the ever-changing scenes, characters, and environmental conditions as the story unfolds.
The environment then presents the model with a sequence of narrative sections derived from key milestones in the play log. 
At each step, the model receives the current scene description and state updates, and outputs the next action or response to advance the interaction. 
Successful progression requires the model to consistently maintain and update observable variables across turns because later events frequently depend on states introduced or modified earlier. 
This setup therefore evaluates whether models can reliably track dynamic surface states during multi-step interactions.


\subsubsection{No Case Should Remain Unsolved}
\textit{No Case Should Remain Unsolved} is a detective adventure visual novel game.
The core gameplay involves piecing together numerous fragmented conversations and clues according to the correct speaker identity and timing, thereby reconstructing the sequence of events and uncovering the truth behind the case.
Specifically, players are required to infer speaker identities and connect relevant clues based on dialogue content and keywords. By searching for in-game items such as passwords and keys, they gradually unlock 54 dialogue fragments and uncover two hidden truths embedded in them. During this process, players encounter multiple types of lock events. Pink locks evaluate evidence matching ability. Purple locks assess the ability to infer and locate specific time points. Yellow locks measure the ability to reconstruct the temporal order of multiple events. The game is completed only when players reorder all dialogues according to the correct timeline.

This game primarily evaluates \textbf{Temporal Associative Memory}, that is, the ability to connect events across different time points. The interaction follows a temporal reconstruction paradigm, where clues are revealed gradually and only become meaningful when linked to earlier or later events in the timeline. To solve the case, the model must remember previously encountered clues, associate them with newly discovered evidence, and organize them into a coherent temporal structure. This setting evaluates whether the model can maintain temporally distributed memory fragments and correctly relate them as the investigation unfolds.

\begin{figure*}[t]
    \centering
    \includegraphics[width=1\linewidth]{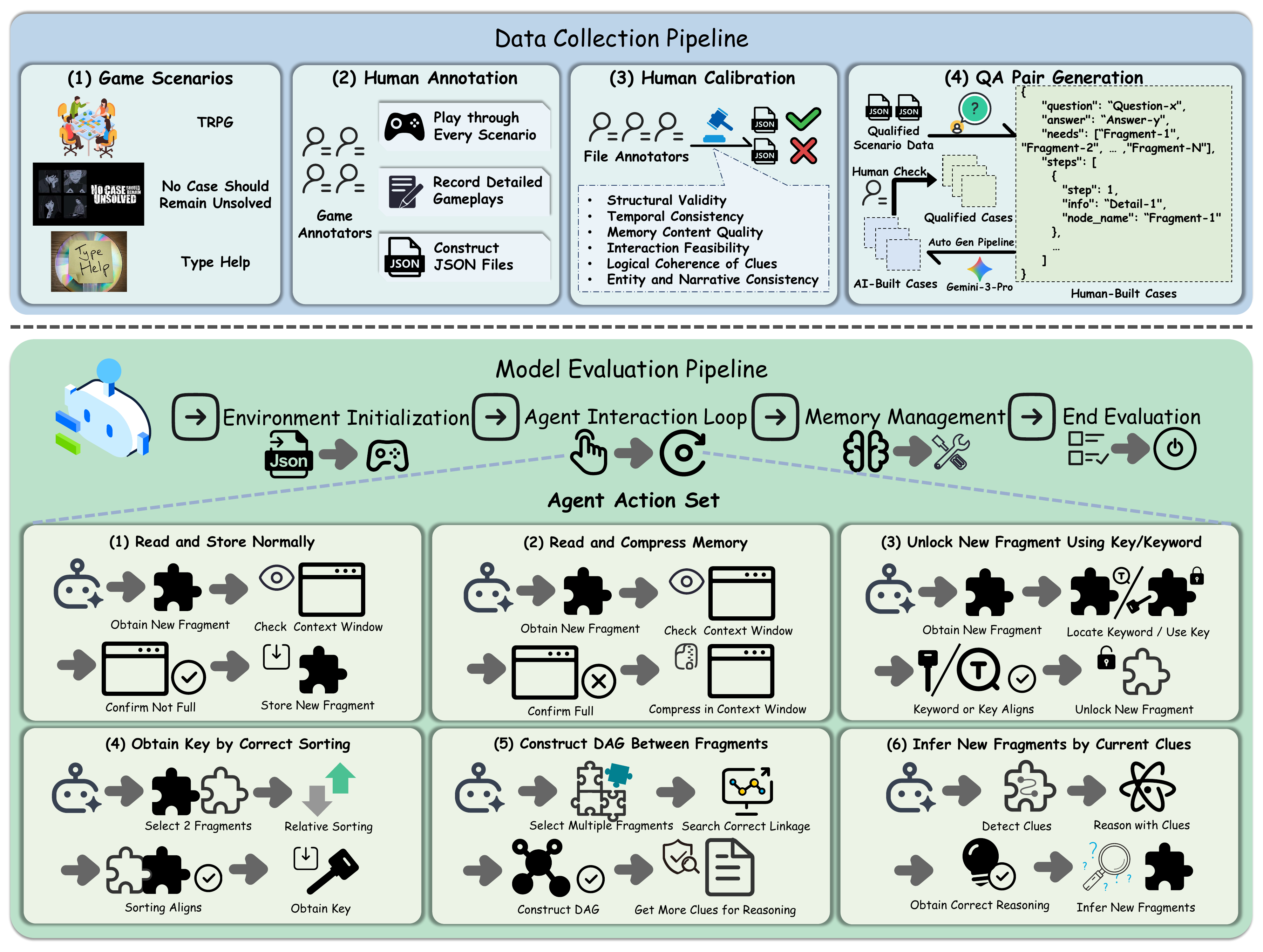}
    \caption{\textbf{Overview of Game Ground.} Game Ground consists of two components, data collection pipeline as well as model evaluation pipeline. } 
    \label{fig:2}
\end{figure*}

\subsubsection{Type Help}
\textit{Type Help} is a web-based, text-only deduction puzzle game.
Players take on the role of an investigator and access character dialogues by inferring and entering the correct file names in the database. Through the unlocked files and newly revealed clues, they gradually deduce identities, relationships, and movement patterns, ultimately reconstructing the full story and unraveling the mystery.

This game primarily evaluates \textbf{Reasoning-Based Memory}, which reflects the ability to perform logical inference over accumulated memory fragments.
Abstractly, the task implements a \emph{memory-to-reasoning} paradigm: the environment hides many small information units whose relevance is only revealed when they are combined and interpreted together as premises for higher-level conclusions. 
At each turn the model receives the current interface state and issues a command. 
Retrieved fragments are added to the model’s memory buffer.
Correct task performance requires actively synthesizing these fragments into intermediate hypotheses (e.g., candidate identities or inferred links), using those hypotheses to guide subsequent retrievals, and ultimately producing a final structured answer. 
Because many clues are ambiguous in isolation, success in this game depends on a model’s ability to (i) accumulate and index relevant fragments, (ii) chain multiple pieces of evidence into logical steps, and (iii) use inferred relations to unlock further evidence.

\subsection{Game Ground Overview}

MemGround consists of two tightly coupled components: (1) a human-grounded data construction pipeline and (2) a unified interactive evaluation framework for LLMs. 
Together, they form a reproducible and extensible \textit{Game Ground} for memory-centric benchmarking, as shown in Figure \ref{fig:2}.

\subsubsection{Data Collection Pipeline}
The construction of MemGround begins with real human gameplay in authentic game environments. For each selected game, trained annotators play through the full storyline under normal player conditions, without artificial simplifications. During gameplay, we record dialogue content, state transitions, unlocked clues, file retrieval paths, temporal relationships, and milestone events.
After gameplay completion, all textual materials, including dialogue fragments, state updates, file contents, unlocking conditions, and action traces, are systematically extracted and structured into standardized JSON files. 
To ensure reliability and structural correctness, another group of annotators performs cross-validation and calibration. This step includes verifying temporal consistency, checking dependency completeness, removing redundant edges, etc. Only after double annotation and reconciliation are the finalized datasets integrated into the evaluation environment.
This two-stage human annotation process guarantees that the benchmark faithfully reflects real human exploration patterns and preserves the intrinsic logical structure of each game scenario.

\subsubsection{Model Evaluation Pipeline}

Model evaluation is conducted through a modular interactive framework, which simulates dynamic gameplay between models and the game environment.
The system architecture is designed to separate environment logic, agent policy, memory mechanisms, and logging utilities, ensuring reproducibility and extensibility.
At a high level, the evaluation process is constructed as:

\paragraph{Environment Initialization}
The environment factory loads a specific game from the structured dataset. The corresponding prompt builder constructs contextual observations based on the current game state.

\paragraph{Agent Interaction Loop}
The agent runner executes a main loop including four parts.
Firstly, the environment produces an observation, such as a newly unlocked fragment or a file content.
Secondly, the agent’s policy module invokes the LLM to generate a decision of retrieval query, ordering adjustment, action selection, or story summarization.
Thirdly, the environment updates internal states according to the decision logic.
Lastly, the memory module stores and optionally compresses interaction history.


\paragraph{Memory Management}
The memory subsystem provides a set of modular tools that support both in-context baselines and agents equipped with external memory. 
These tools operate over a persistent store of interaction histories and environment observations collected during gameplay. 
Each tool exposes a simple input–output interface that enables models or agent frameworks to manage and retrieve memory fragments.

\begin{itemize}

\item \textbf{StoreMemory}: Stores new memory fragments from the current interaction.  

\textit{StoreMemory(interaction, observation)} $\rightarrow$ \textit{memory\_fragment}

\item \textbf{CompressMemory}: Summarizes or compresses stored memory fragments to simulate realistic context window constraints.  

\textit{CompressMemory([memory\_fragment\_1, ..., memory\_fragment\_n])} $\rightarrow$ \textit{[compressed\_memory]}

\item \textbf{RetrieveMemory}: Retrieves relevant past information via keyword matching or vector-based semantic search.  

\textit{RetrieveMemory(query)} $\rightarrow$ \textit{[memory\_fragment\_1, memory\_fragment\_2, ...]}

\item \textbf{IndexMemory}: Indexes stored memory representations using a Faiss-backed vector database for scalable similarity search.  

\textit{IndexMemory(memory\_embeddings)} $\rightarrow$ \textit{vector\_index}

\end{itemize}

\begin{figure*}[t]
    \centering
    \includegraphics[width=0.85\linewidth]{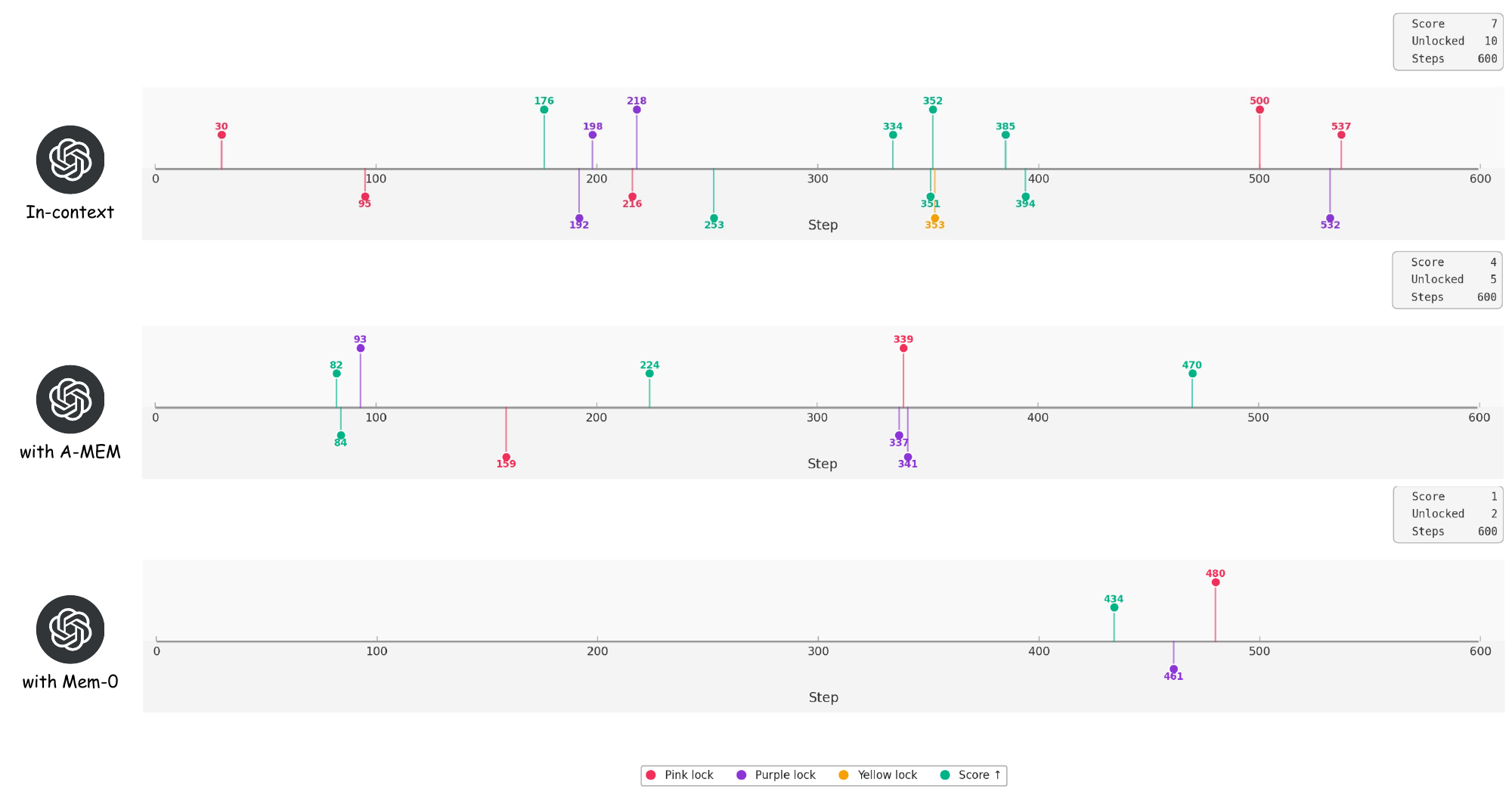}
    \caption{The progress chart of GPT-5.2 and its memory agents in No Case Should Remain Unsolved. The pink, purple, yellow, and green points stand for the model unlocking a pink-lock file, unlocking a purple-lock file, unlocking a yellow-lock file, and sorting 2 memory fragments in correct timeline, respectively.}
    \label{fig:game2}

    \vspace{0.1cm}

    \includegraphics[width=0.8\linewidth]{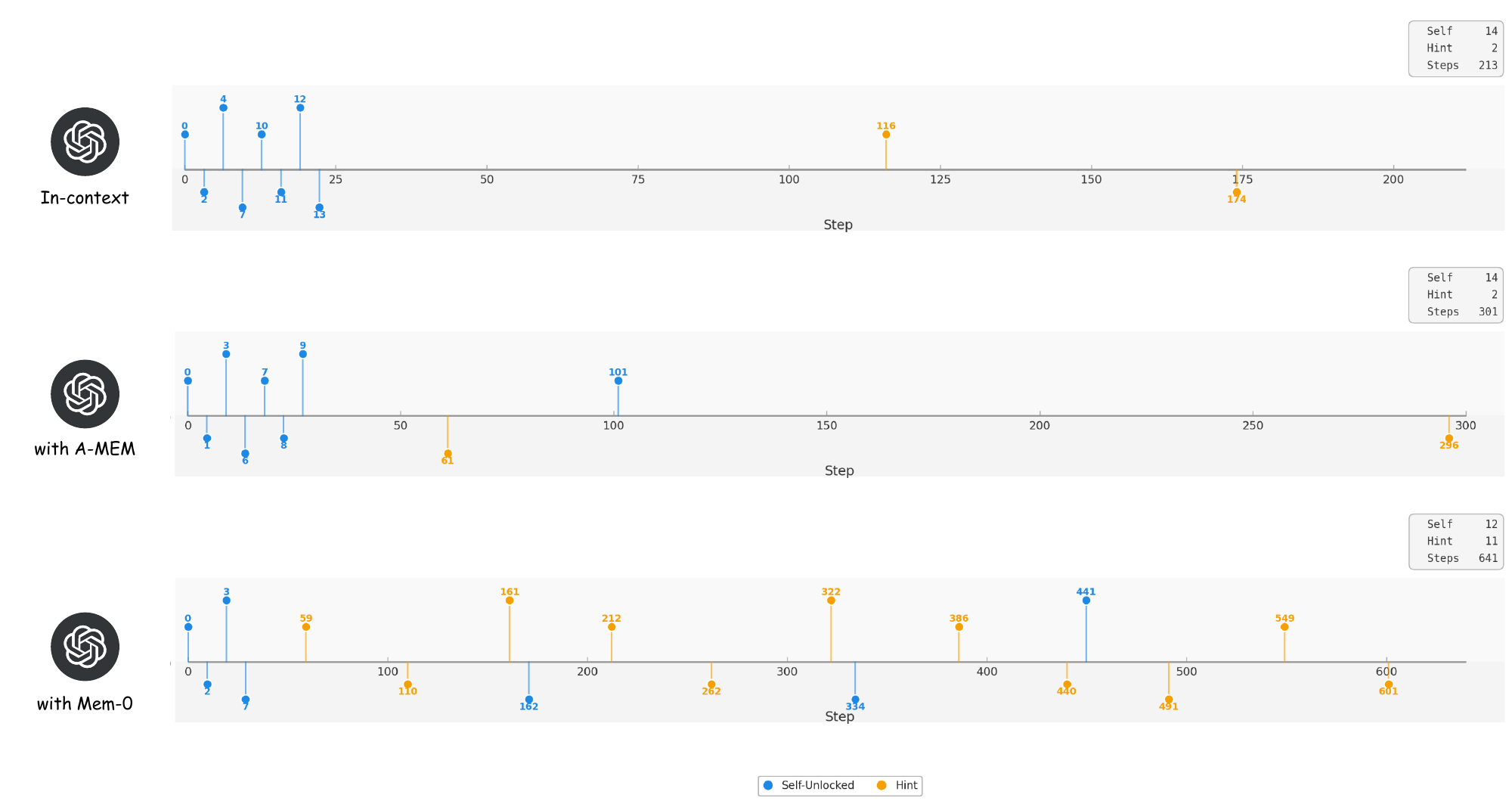}
    \caption{The progress chart of GPT-5.2 and its memory agents in Type Help. The blue and yellow points stand for the model unlocking a file by itself and unlocking a file by receiving a hint, respectively.}
    \label{fig:game3}
\end{figure*}

\section{Experiments}
\begin{table*}[t]
    \centering
    \small
    \setlength{\tabcolsep}{2.5pt}
    \begin{tabular}{llc ccc cc}
        \toprule
        \multicolumn{1}{c}{\multirow{2}{*}{Model}} &
        \multicolumn{1}{c}{\multirow{2}{*}{Method}} 
        & \multicolumn{1}{c}{TRPG} 
        & \multicolumn{3}{c}{No Case Should Remain Unsolved} 
        & \multicolumn{2}{c}{Type Help} \\
        
        \cmidrule(lr){3-3}
        \cmidrule(lr){4-6}
        \cmidrule(lr){7-8}
        
        &
        & QA Overall$\uparrow$ (\%)
        & QA Overall$\uparrow$ (\%) & MFCO$\uparrow$ (\%) & MFU$\uparrow$ (\%)
        & QA Overall$\uparrow$ (\%) & MFU$\uparrow$ (\%)  \\
        \midrule
        
        \rowcolor{gray!20}\multicolumn{8}{c}{\textbf{Closed-Source}} \\
        \midrule
        
        Gemini-3-Pro-Preview & In-context & 26.28 & 41.61 & \highblue{\textbf{45.46}} & \highblue{\textbf{83.33}} & \highgreen{28.62} & \highblue{\textbf{50.91}}
        \\
        \midrule
        
        DeepSeek-V3.2 & In-context & 31.41 & 37.58 & 27.27 & \highgreen{81.48} & 27.98 & 17.27 \\
        \midrule
        
        Claude-Opus-4.6 & In-context & 38.07 & \highblue{\textbf{47.76}} & \highgreen{38.64} & \highgreen{81.48} & 23.81 & \highgreen{37.27} \\
        \midrule

        \multirow{3}{*}{GPT-5.2}
            & In-context & \highgreen{51.51} & 28.56 & 15.91 & \highgreen{81.48} & 23.61 & 14.55   \\
            & Mem0       & 48.06 & 31.87 & 2.27 & 66.67 & \highblue{\textbf{28.65}} & 20.91 \\
            & A-MEM      & \highblue{\textbf{53.79}} & \highgreen{41.72} & 9.09 & 72.22 & 21.80 & 14.55 \\
        
        \midrule
        
        \rowcolor{gray!20}\multicolumn{8}{c}{\textbf{Open-Source}} \\
        \midrule
        
        \multirow{3}{*}{Qwen3-VL-32B-Instruct}
            & In-context & 25.78 & 13.03 & 0.00 & 66.67 &  14.02& 14.55 \\
            & Mem0       & 33.06 & 18.43 & 0.00 & 68.52 & 16.74 & 15.45 \\
            & A-MEM      & 38.71 & 8.59 & 0.00 & 66.67 & 17.27 & 14.55 \\
            
        \bottomrule
    \end{tabular}
    \caption{Main results of multiple state-of-the-art LLMs and memory agents in MemGround. $\uparrow$ indicates higher is better.  The highest and second highest scores are marked in \textcolor{blue}{blue} and \textcolor[RGB]{0,119,51}{green}, respectively. 
    }
    \label{tab:mainresults}
\end{table*}

\subsection{Baselines}
\paragraph{Large Language Models.} We evaluate MemGround with a set of strong state-of-the-art LLMs that represent current frontier capabilities in long-context reasoning and memory.
We include strong closed-source proprietary frontier models, which are GPT-5.2~\citep{singh2025openai}, Gemini-3-Pro-Preview~\citep{team2023gemini}, DeepSeek-V3.2~\citep{liu2025deepseek}, and Claude-Opus-4.6~\citep{anthropic2024claude}, together with the competitive open-weight model Qwen3-32B-Instruct~\citep{yang2025qwen3}.
All baselines are evaluated in an in-context configuration to quantify how effectively they can reconstruct memory using only their context window and internal parametric knowledge.

\paragraph{Memory Agents.} Beyond in-context baselines, we evaluate two representative memory-augmented frameworks Mem0~\citep{chhikara2025mem0} and A-MEM~\citep{xu2025mem}) on the open-weight model Qwen-3-32B-Instruct~\citep{yang2025qwen3} and the representative closed-source model GPT-5.2~\citep{singh2025openai}, to measure the impact of explicit long-term memory modeling in text-based interactive environments.

\subsection{Implementation Details}
In our evaluation, a step is defined as one complete interaction cycle in which the model receives information from the environment, retrieves relevant information, and makes a decision.

In \textit{TRPG}, to reduce the possibility that models have prior knowledge of the rules, we select three game logs from relatively new rule systems.

In \textit{No Case Should Remain Unsolved}, we set the maximum number of interaction steps to 600.

In \textit{Type Help}, we set the maximum number of interaction steps to 1000. We also introduce a hint mechanism: if the model fails 50 consecutive times, the next files is automatically unlocked. To prevent over-reliance on hints, we further apply an early stopping rule that terminates the run if the model does not discover any new files within 200 consecutive steps.

\subsection{Evaluation Metrics}

\paragraph{Question-Answer Score (QA Overall).}
To quantitatively evaluate models’ ability to retrieve and reason over long-term accumulated memory, we construct a set of open-ended multi-hop QA pairs grounded in the intrinsic logic of each game environment.
Formally, \textit{QA Overall} is a composite score derived from a six-dimensional evaluation criterion. For each question, LLM evaluators assess the model's response across six predefined dimensions, and obtain \textbf{the aggregated overall score} based on these assessments.
Details can refer to Appendix \ref{QAjudge} and \ref{QAresults}.


\paragraph{Memory Fragments Unlocked (MFU).}
This metric quantifies the game completion progress and measures the model’s reasoning ability.
We define \textit{MFU} as the proportion of valid memory fragments successfully revealed by the model relative to the total number of unlockable fragments in the environment. 
This metric is applied in both \textit{No Case Should Remain Unsolved} and \textit{Type Help}.

\paragraph{Memory Fragments with Correct Order (MFCO).}
This metric evaluates the model’s ability to recover the correct temporal structure from fragmented memories.
\textit{MFCO} is defined as the proportion of evaluable fragment pairs for which the predicted relative temporal order agrees with the gold timeline. Specifically, a pair is counted as correct if the predicted ordering preserves the ground-truth temporal relationship between the two fragments.
This metric is applied in \textit{No Case Should Remain Unsolved} only.


\paragraph{Exploration Trajectory Diagram of Subsections (ETD).}

To construct the subsection-level directed acyclic graph (DAG), we first manually annotate all inter-file references and logical dependencies in the game database. 
For each run of models, we record the sequence of accessed files and construct a trajectory subgraph \(G' = (V', E') \subseteq G\), where \(V'\) contains the visited nodes and \(E'\) includes the actual traversal transitions. 
Based on this way, the metric evaluates the difference between the model’s game path and the human game path at the structural level, rather than only comparing final outcomes.
This metric is applied in \textit{Type Help} only.

\subsection{Main Results and Analysis}

Based on Table \ref{tab:mainresults}, we observe several findings regarding the memory and reasoning capabilities of the evaluated models.
Further details about QA Overall can be found in Appendix~\ref{QAresults}.


    
    

\paragraph{Performance Gap Between Closed and Open-Source Models.} Frontier closed-source models consistently outperform the competitive open-source baseline. This divide is most striking in temporal structural alignment, where Qwen3 achieves near-zero scores in the MFCO metric across all configurations.

\paragraph{Degradation Across Hierarchical Memory Tasks.} Model performance sharply deteriorates as the cognitive demands of the scenarios increase. While models perform reasonably well in sequential Surface State Memory (TRPG), their QA Overall scores drop significantly in the highly fragmented, Reasoning-Based Memory scenario (Type Help). This highlights that current LLMs excel at passive information extraction but struggle with active, nonlinear synthesis.

\paragraph{Decoupling of Exploration and Structural Reasoning.} High exploration rates do not equate to deep structural understanding. In No Case Should Remain Unsolved, most closed-source in-context models unlock over 80\% of narrative fragments (MFU), yet their MFCO scores remain disproportionately low. This reveals that models can efficiently leverage local cues to find clues, but lack the global reasoning capacity to correctly reconstruct their temporal relationships.

\subsection{Temporal Dynamics of Unlocking}
To further investigate how models manage and utilize memory over extended periods, we analyze the temporal progression of fragment discovery and reasoning across the interactive sessions.
As shown in Figure \ref{fig:game2}, GPT-5.2 display a distinct multi-stage exploration pattern. For instance, it successfully unlocks surface-level files (pink and purple locks) in the early stage. During the middle stage, it correctly sorts the relative order of documents and achieves a score of 7. However, late-stage progress completely stagnates, with no further scores gained. 
In contrast, memory agents demonstrate more sustained and temporally distributed discovery and sorting actions throughout the session. Ultimately, they encounter the same reasoning bottleneck, failing to bridge the final disconnected clues.
In Figure \ref{fig:game3}, models frequently enter extended periods of interaction where no new fragments are unlocked, resulting in long "flat segments" in their progression. These periods indicate that the models are unable to autonomously identify the correct actions or infer the underlying structural dependencies required to move forward. Notably, progress typically resumes only after the framework supplies explicit hints. This alternating cycle of stagnation and hint-driven recovery underscores a critical limitation: despite possessing strong in-context reasoning capabilities, models lack self-directed exploratory persistence. 
Dynamics of other models are in Appendix \ref{ProgessCharts}.



\begin{figure}[t]
    \centering
    \includegraphics[width=1\linewidth]{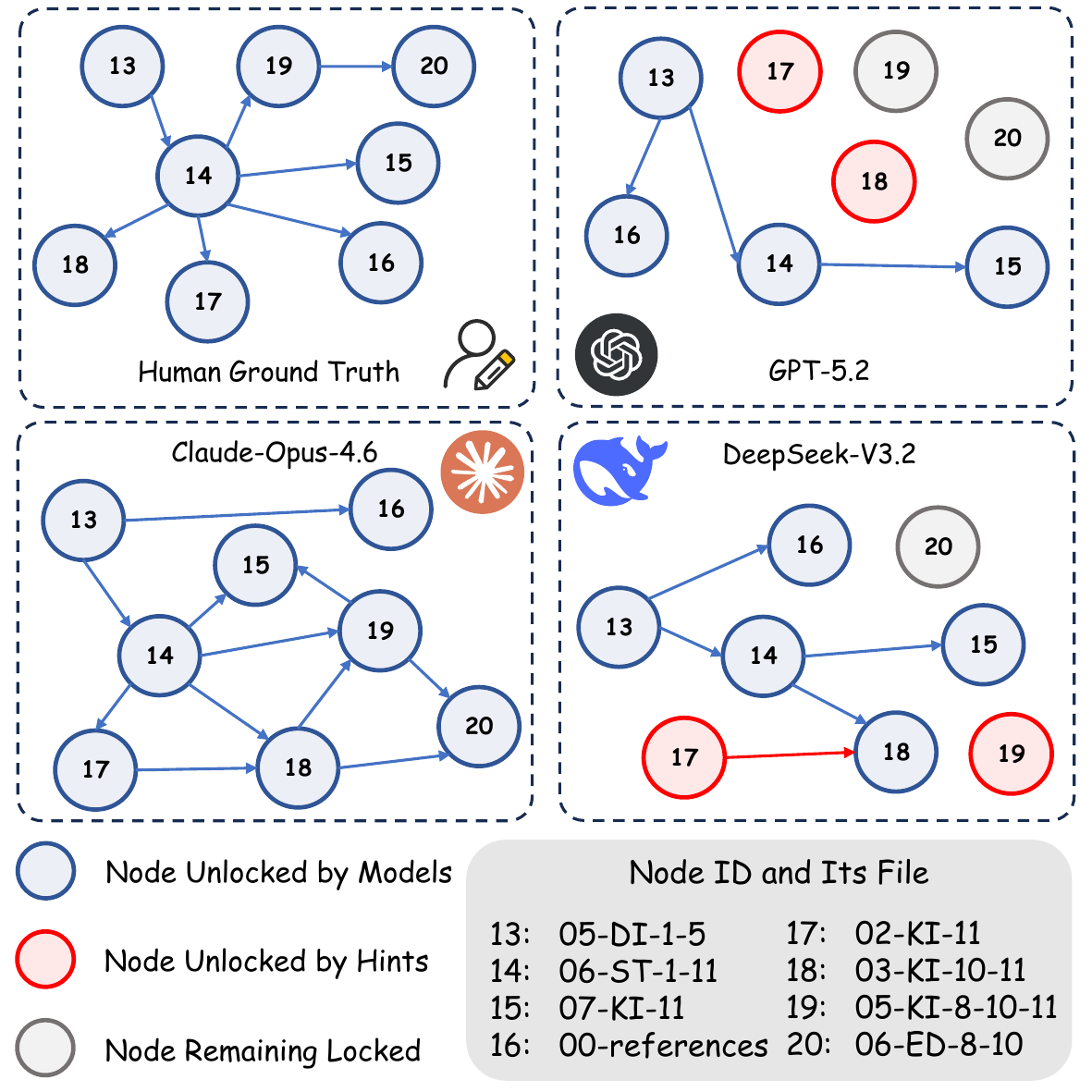}
    \caption{Example ETD of both human and evaluated models in a selected subsection of Type Help.}
    \label{fig:dag}
\end{figure}

\subsection{Comparison of Exploration Trajectory Diagrams (ETD)}
Figure \ref{fig:dag} provides an Exploration Trajectory Diagram (ETD) of both human and evaluated models in a selected subsection of Type Help to compare exploration behaviors at the structural level.
The diagrams contrast the dense, fully connected traversal of the human ground truth against the trajectories generated by Claude-Opus-4.6, GPT-5.2, and DeepSeek-V3.2.
The model graphs illustrate fragmented exploration, highlighting instances where several nodes remain locked (white circles) or are only unlocked via hints (red circles).
The human ground truth displays a dense, fully connected traversal, indicating a logical and comprehensive exploration where all inter-file dependencies are successfully identified and navigated. In contrast, the trajectories of GPT-5.2, Claude-Opus-4.6, and DeepSeek-V3.2 are notably fragmented. The models fail to construct a continuous chain of discovery, often resulting in isolated nodes or broken logical paths.
Also, the models rely on external nudges at the very beginning stage. As indicated by the red-coded nodes in Figure \ref{fig:dag}, a portion of the models' progress is triggered by hints rather than independent deduction.
Gray circles indicate nodes that remained locked throughout the interaction. Notably, even with advanced reasoning, DeepSeek-V3.2 misses key files (such as Node 19) crucial for understanding the complete narrative. This highlights a critical gap between models' fragmented exploration and systematic human-like memory. (See Appendix \ref{DAG}.)

\section{Conclusion}
In this work, we introduce MemGround, a benchmark designed to evaluate long-term memory and reasoning capabilities of LLMs and agents in interactive gamified environments. 
MemGround evaluates hierarchical memory abilities through three progressive and tiered game scenarios, covering Surface State Memory, Temporal Associative Memory, and Reasoning-Based Memory. 
MemGround further proposes multi-dimensional evaluation metrics tailored to the gamified scenarios.
Experiments across multiple models reveal that current systems still struggle to consistently maintain coherent long-term memory then perform reliable reasoning over distributed memory fragments. 

\section*{Ethical Statement}
All data and tasks in MemGround do not contain personally identifiable information or sensitive personal records and are only used for academic research.
Although the benchmark itself does not involve human subjects, improved memory capabilities in AI agents may introduce potential risks in real-world applications, such as unintended long-term storage of user information. 
Individuals using this benchmark should consider responsible memory management strategies, including privacy-aware storage, selective memory updating, and mechanisms for forgetting sensitive information.

\section*{Limitations}

Despite providing a structured benchmark for evaluating hierarchical memory in interactive environments, MemGround has several limitations. 
First, MemGround is constructed from only three representative game scenarios. Although these tasks cover surface state tracking, temporal association, and reasoning over accumulated fragments, they cannot fully represent the diversity of real-world long-term memory demands across broader interactive applications. 
Second, MemGround focuses primarily on text-based interaction environments. Many real-world agent systems operate in multimodal settings involving visual interfaces, spatial navigation, or embodied interaction, which are not captured in the current evaluation framework. 
Future work will expand MemGround toward multimodal interactive environments, integrating visual interfaces, spatial navigation, and embodied interaction settings to better evaluate long-term memory capabilities of agents in real-world scenarios.



\bibliography{custom}





\appendix
\label{sec:appendix}

\newpage


\section{Experimental Setup}

\subsection{Base Prompts}
\subsubsection{Base Prompt in TRPG}

\begin{tcolorbox}[title=Base Prompt in TRPG]
You are a deep reading comprehension expert for TRPG campaign narratives, currently reviewing the dialogue log for the {story}.
Please read carefully and memorize the dialogue content, as you will need to answer questions about the story afterward.

\vspace{3pt} 

Summarize the following TRPG session log into a single cohesive new summary.

\vspace{3pt} 

INPUT STRUCTURE: 
The content may contain two types of material:
\begin{enumerate}[leftmargin=2em, label=--, itemsep=0.5pt]
      \item Paragraphs starting with [Story Summary]: Previously compressed summaries, are already highly condensed. Every sentence contains critical, irreplaceable information.
      \item Raw session lines ([playerX] / [GM] format): Original TRPG records are with lower information density. Redundant content can be trimmed.
\end{enumerate}

\vspace{3pt} 

COMPRESSION TARGET: Merge both types into one coherent narrative. Aim to compress the total input to 50\%-60\% of its original length. Use your own judgment on how to weight each part.

\vspace{3pt} 

HARD CHARACTER LIMIT: Output Must be between \{min\_chars\} and \{max\_chars\} characters. This is a non-negotiable constraint:
\begin{enumerate}[leftmargin=2em, label=--, itemsep=0.5pt]
      \item Do NOT write fewer than \{min\_chars\} characters. Falling short means over-compression, and you must add back detail.
      \item Do NOT exceed \{max\_chars\} characters. Exceeding means insufficient compression, and you must cut further.
\end{enumerate}

\vspace{3pt} 

Preserve all information valuable for understanding the story, including but not limited to:
\end{tcolorbox}

\begin{tcolorbox}[title=]
\begin{enumerate}[leftmargin=2em, label=--, itemsep=0.5pt]
      \item Character actions and movements: what each character did, where they went, who they interacted with, and the outcomes.
      \item Inner monologue and emotions: characters' true thoughts, feelings, and reactions (even brief asides should be kept).
      \item Contradictions between words and actions: what characters said versus what they actually thought.
      \item Skill checks and dice rolls: skill names, results (success/failure/hard success/critical success/critical failure) and their narrative impact.
      \item Sanity and HP changes: specific numerical changes and their triggers.
      \item NPC behavior: important NPCs' words, attitudes, information revealed or concealed.
      \item Clues and foreshadowing: investigation findings, anomalies, unresolved mysteries.
      \item Consequences of events: changes in situation, relationships, or newly gained intelligence.
      \item Scene information: time, location, atmosphere, and environmental details.
\end{enumerate}

\vspace{3pt}

FINAL CHECK: Before outputting, verify your character count is between \{min\_chars\} and \{max\_chars\}. Adjust if needed.

\vspace{3pt}

Output in English. Do not add any prefix or explanation.
\end{tcolorbox}

\subsubsection{Base Prompt in No Case Should Remain Unsolved}

\begin{tcolorbox}[title=Base Prompt in No Case Should Remain Unsolved]
You are a reasoning game agent playing a puzzle game called No Case Should Remain Unsolved.

\vspace{3pt}

Game Objective: \{self.goal\_instruction\}

\vspace{3pt}

Game Mechanics:

\end{tcolorbox}

\begin{tcolorbox}[title=]

\vspace{3pt}

1. Keyword Discovery: When reading event text for the first time, keywords hidden within will be automatically discovered and added to your keyword pool.

\vspace{3pt}

2. Event Unlocking: Use keywords to unlock new events associated with that keyword. After unlocking, you'll know the event name but need to actively read it to get the full content.

\vspace{3pt}

3. Event Reading:
\begin{enumerate}[leftmargin=2em, label=--, itemsep=0.5pt]
      \item Unread Events: Select events from the unread events list for first-time reading to get full content and discover keywords.
      \item Read Events: You can re-read any event from the read events list to review their content.
\end{enumerate}

4. Character Event Ordering: Each event involves multiple characters. You need to infer the chronological order of events from each character's perspective. Submitting correct orderings earns points.

\vspace{3pt}

5. Scoring and Keys: For each correctly ordered event pair (an "earlier-later" relationship from a character's perspective that are consecutive), you earn 1 point. Accumulating a certain score automatically gives you a key. Already scored event pairs won't be scored again.

\vspace{3pt}

6. Lock Mechanism:
\begin{enumerate}[leftmargin=2em, label=--, itemsep=0.5pt]
      \item Pink and Purple locks: Unlock by answering questions.
      \item Yellow lock: Unlock by consuming 1 key.
\end{enumerate}

Strategy Suggestions (in priority order):
\vspace{3pt}

1. Prioritize using keywords to unlock new events: When keywords are available, use them to unlock new events and expand explorable content.
\vspace{3pt}

2. Prioritize reading unread events: Read events from the unread events list to extract keywords and character information.

\vspace{3pt}

3. When no unread events and no keywords, try submitting orderings:
\end{tcolorbox}

\begin{tcolorbox}[title=]

\begin{enumerate}[leftmargin=2em, label=--, itemsep=0.5pt]
      \item Ordering strategy: First determine which character each event belongs to, then after confirming correct event attribution, infer the chronological order within that character's events.
      \item Even if incomplete information, you can try - correct orderings earn points and keys.
\end{enumerate}

4. When you have keys, unlock yellow-locked events: Use keys to unlock important yellow-locked events.
\vspace{3pt}

5. When you have sufficient information, answer questions to unlock pink/purple locks: Infer answers based on read events.
\vspace{3pt}

6. You can select events from the read events list to re-read if you need to review details.

\vspace{5pt}
Important Notes:

\begin{enumerate}[leftmargin=*, label=--, itemsep=0.5pt]
      \item Nodes starting with "talk-" (e.g. "talk-1", "talk-2") contain no important reasoning information, do not participate in character event ordering, do not need to be repeatedly read, and must not influence your judgment on event attribution or ordering.
      \item Hint: There is only one event under the Eden Kindergarten character.
      \item Don't unlock too many locked events too early, prioritize exploring unlocked events.
\end{enumerate}
\end{tcolorbox}

\subsubsection{Base Prompt in Type Help}

\begin{tcolorbox}[title=Base Prompt in Type Help] 
You are a game agent specialized in puzzle-solving games. In this game, you need to obtain information by entering \textbf{filenames} that follow certain specific rules. You need to reasonably infer the file naming rules based on available information, \textbf{deduce possible correct filenames} to unlock more files, and use the information obtained from files to \textbf{reconstruct the entire story}.

\vspace{3pt} 
Strategy:

\end{tcolorbox}

\begin{tcolorbox}[title=]
1. Prioritize opening unlocked but not yet viewed files to help you gain more information.
\vspace{3pt}

2. Question marks in filenames indicate parts you need to guess.
\vspace{3pt}

3. For failed files, \textbf{do not try the same failed filename again}, try other combinations.
\vspace{3pt}

4. When guessing filenames, carefully analyze the naming patterns in unlocked files, such as the meaning of numbers, the ordering relationship of number sizes, the meaning of letters, etc.
\vspace{3pt}

5. For character numbers, do not guess numbers that are too large and haven't appeared in the text information.
\vspace{3pt}

6. \textbf{Focus on character movement information}: Carefully read dialogues and scene descriptions to extract these clues for deducing the next filename:
      \begin{enumerate}[leftmargin=2em, label=--, itemsep=0.5pt]
          \item A character says "I'm going to [location]", "Go find him at [location]", or is summoned to a location. This character will appear in the next time slot's file for that location.
          \item A character leaves the current scene. They will no longer appear in subsequent files of the current location, but will appear in files of the new location.
          \item New characters enter the scene. Their numbers should be added to the current location's subsequent file name.
          \item Multiple characters heading to the same location. All their numbers should appear together in the target location's filename.
       \end{enumerate}

7. \textbf{Pay special attention to the beginning and end of each node's text}: Character movement clues tend to concentrate there. The opening describes who enters the scene or where they came from, while the ending describes who leaves, where they are going, or what action is planned next.
\end{tcolorbox}

\subsection{QA Test Prompts}

\subsubsection{QA Test Prompt in TRPG}

\begin{tcolorbox}[title=QA Test Prompt in TRPG] 

You are an expert in deep reading comprehension of TRPG narrative records.
You have read a TRPG session log in order (earlier content may have been summarized due to context limits).
Based on your memory and current context, respond using the following fixed format (do NOT omit either label):

\vspace{3pt}

Reasoning: [2-4 sentences of analytical reasoning, using logical connectives such as "because... therefore...", "this suggests...", "taken together...", to show your reasoning chain]

\vspace{3pt}

Answer: [1-2 sentence concise conclusion that directly answers the question]

\vspace{3pt}

Answer in English. All content must be grounded in the story.
\end{tcolorbox}

\subsubsection{QA Test Prompt in No Case Should Remain Unsolved}
\begin{tcolorbox}[title=QA Test Prompt in No Case Should Remain Unsolved] 
You are an expert in deep reading comprehension of detective/mystery game records.
You have read the interrogation logs of a case in order (earlier content may have been
summarized due to context limits).
Based on your memory and current context, respond using the following fixed format
(do NOT omit either label):

\vspace{3pt}

Reasoning: [2-4 sentences of analytical reasoning, using logical connectives such as
"because... therefore...", "this suggests...", "taken together...", to show your reasoning chain]

\vspace{3pt}

Answer: [1-2 sentence concise conclusion that directly answers the question]

\vspace{3pt}

Answer in English. All content must be grounded in the story.
\end{tcolorbox}

\subsubsection{QA Test Prompt in Type Help}
\begin{tcolorbox}[title=QA Test Prompt in Type Help] 
You are an expert in deep reading comprehension of mystery puzzle game records.
You have explored a series of files in a mansion investigation game in order (earlier content
may have been summarized due to context limits). Each file reveals information about characters,
locations, and events at specific times.
Based on your memory and current context, respond using the following fixed format
(do NOT omit either label):

\vspace{3pt}

Reasoning: [2-4 sentences of analytical reasoning, using logical connectives such as
"because... therefore...", "this suggests...", "taken together...", to show your reasoning chain]

\vspace{3pt}

Answer: [1-2 sentence concise conclusion that directly answers the question]

\vspace{3pt}

Answer in English. All content must be grounded in the files you have read.
\end{tcolorbox}

\subsection{Judge Prompts for QA Test}\label{QAjudge}
Since our QA pairs are open-ended, the separate LLM (Gemini-3.1-Pro-Preview) is required for judge scoring. 
The scoring criteria are primarily divided into the four dimensions, which are answer consistency (Acc Judge), evidence grounding assessment (Cit Judge), format compliance assessment (Inst Judge), and reading comprehension and evidence coverage (Read Judge). 

\subsubsection{Prompt of Acc Judge}
\begin{tcolorbox}[title=Prompt of Acc Judge]
You are an objective and strict answer consistency evaluator.
Determine whether the "predicted answer" and the "gold answer" are semantically equivalent. 
\vspace{3pt}

Rules:
\begin{enumerate}[leftmargin=*, label=--, itemsep=0.5pt]
    \item CONSISTENT: Both convey the same core meaning (different wording but referring to the same entity/relation/fact).
    \item INCONSISTENT: They point to different entities or contain substantive contradictions.
\end{enumerate}

\end{tcolorbox}

\begin{tcolorbox}[title =]
\begin{enumerate}[leftmargin=*, label=--, itemsep=0.5pt]
    \item UNDETERMINABLE: The gold answer itself is ambiguous, or consistency cannot be determined.
\end{enumerate}

Note TRPG data specifics: player labels (playerX) are interchangeable with in-game character names; if the core fact matches, it is CONSISTENT.
\vspace{3pt}

Output ONLY JSON, nothing else:
\begin{verbatim}
{
    "result": "CONSISTENT"
            |"INCONSISTENT"
            |"UNDETERMINABLE", 
    "reason": "brief reason 
            (less than 30 words)"
}
\end{verbatim}
\end{tcolorbox}

\subsubsection{Prompt of Cit Judge}
\begin{tcolorbox}[title=Prompt of Cit Judge]
You are an objective evidence-grounding evaluator.
Given the key evidence IDs for a question and the model's predicted answer, assess whether the answer correctly utilizes the information from those key story moments.

\vspace{3pt}

Rules:
\begin{enumerate}[leftmargin=*, label=--, itemsep=0.5pt]
    \item HIGH: The predicted answer clearly reflects the information from the key events/dialogues indicated by the evidence (explicit ID citation is NOT required).
    \item MEDIUM: Partially reflects the evidence — uses some relevant information but misses major evidence points.
    \item LOW: Does not reflect the evidence, is based on incorrect information, or is unrelated to the evidenced events.
\end{enumerate}

Note: Evaluate content grounding, not explicit citation of IDs.

\vspace{3pt}

Output ONLY JSON, nothing else:
\begin{verbatim}
{
    "cit_score": "HIGH"|"MEDIUM"
                |"LOW", 
    "cit_reason": "brief reason 
                (less than 30 words)"
}
\end{verbatim}
\end{tcolorbox}

\subsubsection{Prompt of Inst Judge}
\begin{tcolorbox}[title=Prompt of Inst Judge]
You are a strict format and reasoning quality evaluator.
Assess whether the model's answer meets the following instruction requirements:
\begin{enumerate}[leftmargin=*, label=\arabic*, itemsep=0.5pt]
    \item Contains explicit reasoning with logical connectives (e.g., "because", "therefore", "this suggests", "taken together", "thus", "hence", "which means").
    \item Has a clear structure with a distinct reasoning part and a conclusion.
    \item Appropriate length (not a single sentence, not excessively long).
\end{enumerate}

Judgment:
\begin{enumerate}[leftmargin=*, label=--, itemsep=0.5pt]
    \item PASS: Fully satisfies all three criteria.
    \item FAIL: Fails to meet any one criterion.
\end{enumerate}

Output ONLY JSON, nothing else:
\begin{verbatim}
{
    "inst_score": "PASS"|"FAIL", 
    "inst_reason": "brief reason 
                (less than 20 words)"
}
\end{verbatim}
\end{tcolorbox}

\subsubsection{Prompt of Read Judge}
\begin{tcolorbox}[title=Prompt of Read Judge]
You are a strict reading comprehension and evidence grounding evaluator.
You are given a question, the model's predicted answer (including reasoning), and the actual text of key evidence passages.
Judge whether the model's reasoning/answer genuinely uses the key information from the provided evidence.

\vspace{3pt}

Rules:
\begin{enumerate}[leftmargin=*, label=--, itemsep=0.5pt]
    \item HIGH: The reasoning or answer clearly reflects core events, dialogues, or details from the evidence text (different wording is fine).
    \item MEDIUM: Partially reflects the evidence — mentions some related information but misses key evidence content.
\end{enumerate}

\end{tcolorbox}

\begin{tcolorbox}[title=]
\begin{enumerate}[leftmargin=*, label=--, itemsep=0.5pt]
    \item LOW: Largely ignores the evidence; the answer is based on guesses or unrelated information.
\end{enumerate}

Note: Verbatim quotation is NOT required. Judge whether the model semantically "read and understood" the evidence.

\vspace{3pt}

Output ONLY JSON, nothing else:
\begin{verbatim}
{
    "read_score": "HIGH"
                |"MEDIUM"
                |"LOW", 
    "read_reason": "brief reason 
                (less than 30 words)"
}
\end{verbatim}
\end{tcolorbox}

\subsection{JSON Environment Examples}

\subsubsection{JSON Environment Example in TRPG}
Structured game scenarios are like:
\begin{lstlisting}
{
  "section": "2. Tracking Spring Energy: First Encounter and Probing",
  "description": "The group follows the Spring Energy light points on their journey, encountering their first battle along the way. Each character demonstrates their abilities, and the beginnings of teamwork emerge.",
  "conversation": [
    {
      "speaker": "player1",
      "text": "Looks like my ability isn't strong enough to locate something that far away."
    },
    {
      "speaker": "player5",
      "text": "\"Sorry, maybe we'll just have to walk there first.\""
    },
    {
      "speaker": "player4",
      "text": "\"No problem, I know the way pretty well, and we can even buy some snacks along the way.\""
    },
    {
      "speaker": "player5",
      "text": "\"Then let's set off.\""
    },
    {
      "speaker": "player4",
      "text": "#Walk ahead to lead the way, while also taking a look around at the surroundings."
    },
    ...
    {
      "speaker": "player1",
      "text": "\"How about planting the grove right here?\""
    },
    {
      "speaker": "player3",
      "text": "#A sudden blow to the head!"
    }
  ],
  "start_ts": "2025-10-24 23:59:53",
  "end_ts": "2025-10-25 00:41:17"
}
\end{lstlisting}

QA pairs are like:
\begin{lstlisting}[]
{
    "question": "In the ritual scene at the end of this Section, what narrative consequences and resource losses resulted from the failed skill checks of Player6 and Player4?",
    "answer": "Player6's Occult check failure (64/55) caused black mist to erupt from the water's surface without showing an image, followed by Player4's check failure (95/57). These two consecutive failures not only narratively caused the ritual bowl to shatter but also mechanically exhausted the feather, wax, and other auxiliary materials, forcing the team to postpone action to find a new container and materials.",
    "evidence": [
      "D01:62",
      "D01:64",
      "D01:67",
      "D01:70",
      "D01:72"
    ],
    "category": 5,
    "id": 4,
    "depth": 1
}
\end{lstlisting}

\subsubsection{JSON Environment Example in No Case Should Remain Unsolved}
Structured game scenarios are like:
\begin{lstlisting}[]
{
  "game": {
    "nodes": [
      {
        "name": "start",
        "sub_name": "start",
        "id": 1,
        "time_id": 1,
        "memory": {
          "description": "The narrator looks back on over seventy years of life through the metaphor of a “jellyfish,” expressing regret for a life spent as a police officer yet failing to save others, and decides to end this life. In a state of mental confusion, Jeon Gyeong talks with the spirit who calls himself Choi Siwon and a Judge, and it is revealed that he failed to find Siwon and caused the case to become a cold case.",
          "key_info": [
            "The narrator compares themself to a “jellyfish,” describing a life drifting with the current and unintentionally hurting those around them.",
            "The narrator reflects on more than seventy years of police work, yet believes they never truly saved anyone, nor offered comfort to families who lost children.",
            "The police station is described as a place filled with sorrow and pain, a place the narrator chose to cling to but that could not bring hope.",
            "The narrator questions their original intention of becoming a police officer, believing the choice came from selfishness and a desire to belong.",
            "The narrator clearly expresses a desire to end a life full of regret, seeing it as the first decision that benefits the world.",
            "The narrator apologizes to others and leaves with unfinished words as a farewell.",
            "An unknown voice initially calls Jeon Gyeong “Grandma” and “Grandma Jeon Gyeong,” causing confusion.",
            "The other party claims to be Choi Siwon and confirms their identity as Siwon’s spirit.",
            "Siwon’s spirit mentions that Siwon’s family appears to have regained peace on the surface.",
            "Jeon Gyeong clearly states that their memory is damaged and they cannot confirm the other party’s identity.",
            "Jeon Gyeong points out that the other party is wearing a police uniform, which does not match their understanding of a “spirit.”",
            "The other party’s identity shifts to a “Judge,” denying they came to judge Jeon Gyeong, but to tell him about his life experiences.",
            "Jeon Gyeong admits that he failed to find Siwon.",
            "Jeon Gyeong confirms that the case, as Siwon’s father Choi Donggeon wished, ultimately became a cold case.",
            "The cold-case outcome means Siwon’s mother will never be able to find Siwon."
          ],
          "characters": [
            {
              "name": "",
              "tag": "",
              "description": "",
              "key_info": [],
              "dialogue": [
                {
                  "name": "What do you mean, a cold case?",
                  "tag": [
                    "playground",
                    "cold case"
                  ],
                  "status": "pending"
                }
              ]
            }
          ],
          "emphasize": {
            "red": [
              "Siwon’s father, Choi Donggeon"
            ]
          }
        },
        "links": [
          {
            "target": "step_1",
            "condition": "auto",
            "recall_info": []
          }
        ]
      },
      {
        "name": "step_1",
        "sub_name": "What do you mean, a cold case?",
        "emphasize": [
          "playground",
          "cold case"
        ],
        "tag": [
          "playground",
          "cold case"
        ],
        "id": 3,
        "time_id": 3,
        "memory": {
          "description": "In the conversation with the police, the respondent to “What do you mean, a cold case?” firmly requests that Siwon’s disappearance be classified as a cold case and asks the police to stop tracking Siwon’s whereabouts. This dialogue shows the respondent’s attitude toward the case and expresses a strong desire for the case to remain a cold case forever.",
          "key_info": [
            "The node content is “What do you mean, a cold case?”",
            "The question is “What do you mean? A cold case? No, Siwon’s dad, we can’t do that,” and the respondent expresses determination and a hope regarding the disappearance: “Please, don’t look for Siwon anymore.”",
            "The respondent clearly points out that Siwon’s disappearance has already been filed and registered as a missing child, stating that the case should be treated as a perfect disappearance case, with no witnesses and no suspects.",
            "The respondent strongly asks the police not to keep pressing for Siwon’s whereabouts, hoping Siwon can live forever in a place her mother can never find.",
            "The respondent kneels down and begs the police to classify Siwon’s disappearance as a cold case, firmly demanding that the investigation not continue."
          ],
          "characters": [
            {
              "name": "Choi Donggeon",
              "tag": "Siwon’s father",
              "description": "Currently marked as the respondent for “What do you mean, a cold case?” (the assigned narrator may change later due to ordering adjustments).",
              "key_info": [
                "In the response to “What do you mean, a cold case?”, he expresses gratitude to the police and asks them to classify the case as a cold case.",
                "He relays Siwon’s view of her mother and hopes the case becomes a perfect disappearance."
              ],
              "dialogue": []
            }
          ]
        },
        "links": [
          {
            "target": "talk-2",
            "condition": "auto",
            "recall_info": []
          }
        ]
      },
      ...
    ]
  }
}
\end{lstlisting}

Specific dialogs for relative sorting are like:
\begin{lstlisting}[]
{
    "text":[
        {
            "name":"Looks like you got mom",
            "time":"2012.2.5.20:15",
            "text":"Police: It seems you received your mother's call.\nPolice: She should have told you everything, right? Are you going to come to the police station with her?\nRespondent: He has never believed in God.\nRespondent: He said that no matter how deep you dig, a shovel can never smell the scent of the soil.\nRespondent: When the child was so seriously ill before, he never prayed for her even once.\nRespondent: So he is simply unwilling to believe.\nRespondent: Thanks to my day and night prayers, we received God's grace.\nRespondent: He always gets angry at himself for not having the ability to change all of this.\nRespondent: Did he really... turn himself in?\nRespondent: If it really was him who took her, it might be because of the child support issue. Since the middle of last year, he mentioned it to me several times.\nRespondent: He said he was short on money and truly couldn’t come up with the child support.\nRespondent: I told him this stage is crucial for the child. I told him to pull himself together as soon as possible. No matter what, Xiyuan is still his child.\nRespondent: But he just wouldn’t pray,\nRespondent: and still wanted to rely on his own strength to change everything."
        },
        {
            "name":"You called Siwon’s dad",
            "time":"2012.2.5.13:28",
            "text":"Police: Did you call Xiyuan’s father?\nRespondent: I’ve already called him several times, but he still hasn’t answered.\nRespondent: We only see each other once a year on Xiyuan’s birthday. We basically don’t keep in touch otherwise.\nRespondent: Honestly, it’s one thing not to answer calls usually, but how can he not answer my call at a time like this!\nRespondent: Actually, on the day the police came to our home, I sent him a text message.\nRespondent: It was the first time the child told me she missed her dad.\nRespondent: Then I told him that Xiyuan missed him and asked if he could come see the child.\nRespondent: But how could he be so cold-hearted..."
        },
        ...
    ]
}
\end{lstlisting}

QA pairs are like:
\begin{lstlisting}[]
{
    "id": 23,
    "question": "After Choi Donggeon's wife passed away, what behaviors or emotional anchors did Myeong-ho and Siwon each exhibit when facing the loss of their mother?",
    "answer": "Siwon became emotionally attached to the nanny, Shin Hyejin, and always called her Mom. Myeong-ho, on the other hand, honored his mother's memory by carrying an umbrella every time he went out, because she had promised him before she passed away that if he ever forgot his umbrella, she would make the rain stop.",
    "needs": [
      "Ms. Shin Hui-jing is",
      "Myeongho also and Siwon",
      "You mean the umbrella"
    ],
    "steps": [
      {
        "step": 1,
        "info": "In [Ms. Shin Hui-jing is], Choi Donggeon mentioned that after his wife passed away, Siwon was very close to the nanny, Shin Hyejin, and always called her mom.",
        "node_name": "Ms. Shin Hui-jing is"
      },
      {
        "step": 2,
        "info": "In [Myeongho also and Siwon], Choi Donggeon mentioned that Myeong-ho overcame the pain of losing his family in his own way, holding onto an item every time he went out.",
        "node_name": "Myeongho also and Siwon"
      },
      {
        "step": 3,
        "answer": "In [You mean the umbrella], Choi Donggeon explains that Myeong-ho always takes an umbrella when going out because of a promise his late wife made to Myeong-ho: if he forgot his umbrella on a rainy day, his mom would make the rain stop. Overall, Siwon placed his emotional attachment on the nanny, while Myeong-ho honors his mother's memory by carrying an umbrella.",
        "node_name": "You mean the umbrella"
      }
    ],
    "character": "Choi Donggeon",
    "char_tag": "Siwon’s father"
}
\end{lstlisting}

\subsubsection{JSON Environment Example in Type Help}
Structured game scenarios are like:
\begin{lstlisting}[]
{
  "game": {
    "nodes": [
      {
        "name": "Start",
        "id": 1,
        "time_id": -2,
        "in_degree": [],
        "links": [
          {
            "target": "Background",
            "condition": "Click to start"
          },
          {
            "target": "Box0",
            "condition": "Click (skip opening)"
          }
        ]
      },
      {
        "name": "Background",
        "id": 2,
        "time_id": -1,
        "in_degree": [
          "Start"
        ],
        "memory": {
          "description": "On Sunday, March 7, 1936, a death occurred at the Galley Villa. The player must investigate the cause of death.",
          "key_info": [
            "On March 7, 1936, a group of locals walked along muddy roads to the Galley Villa and discovered a body. The corpse lay on the front steps, blood flowing, a shocking sight. There were multiple bodies in the villa, with causes of death unknown. Only two bodies were fresh; the causes of death of the others could not be determined. After an official investigation, no further information was made public, and the incident gradually faded from memory. Following the event, aside from limited information about some of the bodies, there was almost no further media coverage."
          ],
          "time": "March 7, 1936",
          "characters": [],
          "links": [
            {
              "target": "Box0",
              "condition": "Click to start game",
              "recall_info": {}
            }
          ]
        }
      },
      ...
    ]
  }
}
\end{lstlisting}

QA pairs are like:
\begin{lstlisting}
{
    "id": 4,
    "time": 5,
    "question": "What item did the visitor, who was guided to the villa by Harry Thornton, find near the body?",
    "answer": "Chapel key",
    "needs": [
      "01-QU-1-11",
      "03-LI-1-4-5-6-7-8-9",
      "04-ST-1-5-8",
      "05-DI-1-5"
    ],
    "tid_max": 4,
    "tid_min": 2,
    "steps": [
      {
        "step": 1,
        "info": "In 01-QU-1-11, Harry Thornton (No. 11) guides John Hobbes to the villa at Quail Lane, confirming John Hobbes as the visitor guided by Harry. John Hobbes's movements after entering the villa must be tracked.",
        "node_name": "01-QU-1-11",
        "tid": 2
      },
      {
        "step": 2,
        "info": "In 03-LI-1-4-5-6-7-8-9, Martha Galley arranges for John Hobbes to spend the night in the Study. Having confirmed that John Hobbes was settled in the Study, we need to backtrack the events that occurred in the Study.",
        "node_name": "03-LI-1-4-5-6-7-8-9",
        "tid": 2
      },
      {
        "step": 3,
        "info": "In 04-ST-1-5-8, an unknown male body is discovered in the Study. John Hobbes is present and apologizes for breaking a wine bottle; John Hobbes finds an item near the body, and the specific details of this item need to be confirmed.",
        "node_name": "04-ST-1-5-8",
        "tid": 3
      },
      {
        "step": 4,
        "answer": "In 05-DI-1-5, John Hobbes hands the key found near the body to Martha Galley for identification. Martha Galley confirms that it is the Chapel key, therefore the answer is the Chapel key.",
        "node_name": "05-DI-1-5",
        "tid": 4
      }
    ]
}
\end{lstlisting}

\begin{table*}[t]
\centering
\small
\setlength{\tabcolsep}{2pt}
\begin{tabular}{llcccccccc}
\toprule
Model & Method & Cam
& Overall$\uparrow$ (\%) & Acc.$\uparrow$ (\%) & Inst.$\uparrow$ (\%) 
& Cit.$\uparrow$ (\%) & Read.$\uparrow$ (\%)
& Comp.$\uparrow$ (\%) & Dep.$\uparrow$ (\%) \\
\midrule

\rowcolor{gray!20}\multicolumn{10}{c}{\textbf{Closed-Source}} \\
\midrule

\multirow{3}{*}{Gemini-3-Pro-Preview}
& \multirow{3}{*}{In-context}
& T & 23.66 & 21.39 & 97.86 & 42.78 & 15.78 & 16.88 & 19.66 \\
& & S & 30.81 & 32.06 & 96.18 & 70.23 & 18.32 & 20.00 & 29.17 \\
& & C & 24.38 & 21.98 & 90.11 & 57.14 & 15.38 & 17.50 & 21.24 \\
\midrule

\multirow{3}{*}{DeepSeek-V3.2}
& \multirow{3}{*}{In-context}
& T & 27.48 & 26.74 & 96.79 & 60.43 & 12.30 & 25.97 & 26.75 \\
& & S & 33.14 & 32.82 & 97.71 & 70.23 & 19.08 & 29.09 & 32.08 \\
& & C & 33.60 & 32.97 & 100.00 & 63.74 & 19.23 & 32.50 & 32.54 \\
\midrule

\multirow{3}{*}{Claude-Opus-4.6}
& \multirow{3}{*}{In-context}
& T & 25.60 & 26.74 & 87.70 & 48.13 & 13.37 & 19.48 & 25.71 \\
& & S & 44.77 & 48.85 & 90.84 & 69.47 & 29.39 & 41.82 & 46.71 \\
& & C & 43.84 & 46.15 & 89.01 & 62.64 & 33.52 & 37.50 & 44.87 \\
\midrule

\multirow{9}{*}{GPT-5.2}
& \multirow{3}{*}{In-context}
& T & 36.95 & 41.18 & 93.05 & 96.79 & 14.75 & 31.17 & 40.03 \\
& & S & 55.13 & 58.02 & 100.00 & 97.71 & 38.55 & 52.73 & 55.77 \\
& & C & 62.45 & 65.93 & 100.00 & 95.60 & 50.55 & 55.00 & 63.16 \\

& \multirow{3}{*}{Mem0}
& T & 47.70 & 50.27 & 100.00 & 97.33 & 31.28 & 40.26 & 48.85 \\
& & S & 42.36 & 43.51 & 100.00 & 97.71 & 24.81 & 38.18 & 41.70 \\
& & C & 54.13 & 57.14 & 100.00 & 98.90 & 41.21 & 42.50 & 55.34 \\

& \multirow{3}{*}{A-MEM}
& T & 53.49 & 55.61 & 99.47 & 98.93 & 41.18 & 42.86 & 54.05 \\
& & S & 48.60 & 51.91 & 98.47 & 96.18 & 32.06 & 41.82 & 49.36 \\
& & C & 59.28 & 59.34 & 100.00 & 97.80 & 48.90 & 55.00 & 57.78 \\

\midrule

\rowcolor{gray!20}\multicolumn{10}{c}{\textbf{Open-Source}} \\
\midrule

\multirow{9}{*}{Qwen3-VL-32B-Instruct}
& \multirow{3}{*}{In-context}
& T & 24.57 & 26.20 & 18.18 & 24.94 & 73.80 & 78.07 & 8.82 \\
& & S & 20.59 & 18.32 & 85.50 & 64.89 & 9.16 & 14.55 & 17.61 \\
& & C & 32.17 & 32.97 & 74.73 & 83.52 & 14.29 & 35.00 & 32.19 \\

& \multirow{3}{*}{Mem0}
& T & 34.32 & 36.36 & 85.03 & 78.07 & 20.05 & 27.27 & 34.35 \\
& & S & 25.19 & 22.90 & 90.08 & 70.99 & 13.36 & 20.00 & 21.71 \\
& & C & 39.67 & 38.46 & 85.71 & 73.63 & 29.67 & 37.50 & 37.57 \\

& \multirow{3}{*}{A-MEM}
& T & 39.67 & 40.64 & 81.82 & 75.40 & 24.87 & 31.17 & 38.45 \\
& & S & 30.37 & 29.77 & 87.02 & 80.92 & 22.14 & 14.55 & 28.12 \\
& & C & 46.10 & 50.55 & 81.32 & 79.12 & 31.32 & 42.50 & 47.59 \\

\bottomrule
\end{tabular}
\caption{QA results in TRPG. The column \textit{Cam} stands for Campaign of TRPG. In the \textit{Cam} column, \textit{T}, \textit{S}, and \textit{C} stands for Terror on the Orient Express, Spring Snow Incident, and Cold Wind Howling, respectively. The columns \textit{Acc.}, \textit{Inst.}, \textit{Cit.}, \textit{Read.}, \textit{Comp.}, and \textit{Dep.} stand for Acc Judge score, Inst Judge score, Cit Judge score, Read Judge score, Comprehension score, and Depth score, respectively. $\uparrow$ indicates higher is better. }
\label{tab:qaTRPG}
\end{table*}

\section{More Experimental Results}
This section mainly include QA results, progress charts, and example ETD evaluations.

\subsection{QA Results}\label{QAresults}

In the QA evaluation process, we apply Acc Judge score (Acc.), Inst Judge score (Inst.), Cit Judge score (Cit.), Read Judge score (Read.), Comprehension score (Comp.), and Depth score (Dep.) to serve as 6 dimension for judging the models' answer. We then define the Overall score as the weighted sum of these six dimensions:
\begin{equation}
\begin{aligned}
\text{Overall} =\;& \text{Acc.}\times 30\% + \text{Read.}\times 30\% \\
& + \text{Comp.}\times 15\% + \text{Dep.}\times 15\% \\
& + \text{Inst.}\times 5\% + \text{Cit.}\times 5\%
\end{aligned}
\end{equation}
Here, \textit{Comp.} measures the model's ability to correctly aggregate and reason over information across multiple sections, while \textit{Dep.} reflects the depth of reasoning required, with higher scores assigned to answers involving longer or more complex multi-step inference chains. Descriptions about other judges are already illustrated in Appendix \ref{QAjudge}.

\subsubsection{QA Results in TRPG}
In the TRPG scenario, we collect data from three campaigns, which are Terror on the Orient Express, Spring Snow Incident, and Cold Wind Howling. For each campaign, we construct a set of open-ended QA pairs based on the interaction logs, consisting of 187, 131, and 91 questions for Terror on the Orient Express, Spring Snow Incident, and Cold Wind Howling, respectively.  The results across the three campaigns are reported in Table~\ref{tab:qaTRPG}.
Final \textit{QA Overall} for each model in this game scenario is calculated as the average of the Overall scores across the three campaigns.

Overall, closed-source models consistently outperform open-source models across most metrics, indicating stronger capabilities in maintaining and utilizing long-term memory during interactive gameplay. 
Among all models, GPT-5.2 achieves the best overall performance on the three campaigns, with 36.95 of Terror on the Orient Express, 55.13 of Spring Snow Incident, and 62.45 of Cold Wind Howling, respectively.
We also observe that the difficulty varies across three campaigns. 
Spring Snow Incident and Cold Wind Howling generally yield higher scores on all six-dimension metrics compared to Terror on the Orient Express, suggesting Terror on the Orient Express is more challenging overall.

\begin{table*}[t]
\centering
\small
\setlength{\tabcolsep}{2pt}
\begin{tabular}{llccccccc}
\toprule
Model & Method
& Overall$\uparrow$ (\%) & Acc.$\uparrow$ (\%) & Inst.$\uparrow$ (\%) & Cit.$\uparrow$ (\%) & Read.$\uparrow$ (\%)
& Comp.$\uparrow$ (\%) & Dep.$\uparrow$ (\%) \\
\midrule

\rowcolor{gray!20}\multicolumn{9}{c}{\textbf{Closed-Source}} \\
\midrule

Gemini-3-Pro-Preview & In-context & 41.61 & 33.33 & 96.30 & \textbf{96.30} & 40.74 & \textbf{42.86} & 22.22 \\
\midrule

DeepSeek-V3.2 & In-context & 37.58 & 29.63 & 81.48 & 85.19 & 44.44 & 35.71 & 11.11 \\
\midrule

Claude-Opus-4.6 & In-context & \textbf{47.76} & \textbf{44.44} & 81.48 & 94.44 & \textbf{59.26} & 35.71 & 16.67 \\
\midrule

\multirow{3}{*}{GPT-5.2} & In-context & 28.56 & 18.52 & 92.59 & 92.59 & 35.19 & 14.29 & 6.94 \\
 & Mem0 & 31.87 & 22.22 & \textbf{100.00} & 81.48 & 25.93 & 28.57 & 27.08 \\
 & A-MEM & 41.72 & 40.74 & 96.30 & \textbf{96.30} & 33.33 & \textbf{42.86} & \textbf{22.91} \\

\midrule

\rowcolor{gray!20}\multicolumn{9}{c}{\textbf{Open-Source}} \\
\midrule

\multirow{3}{*}{Qwen3-VL-32B-Instruct} & In-context & 13.03 & 7.41 & 92.59 & 55.56 & 9.26 & 0.00 & 4.17 \\
 & Mem0 & 18.43 & 14.81 & 85.19 & 48.15 & 16.67 & 7.14 & 8.34 \\
 & A-MEM & 8.59 & 7.41 & 70.37 & 33.33 & 1.85 & 0.00 & 4.17 \\

\bottomrule
\end{tabular}
\caption{QA results in No Case Should Remain Unsolved. The columns \textit{Acc.}, \textit{Inst.}, \textit{Cit.}, \textit{Read.}, \textit{Comp.}, and \textit{Dep.} stand for Acc Judge score, Inst Judge score, Cit Judge score, Read Judge score, Comprehension score, and Depth score, respectively. $\uparrow$ indicates higher is better. The highest scores are marked in \textbf{bold}.}
\label{tab:qaDUST}
\end{table*}

\begin{table*}[t]
\centering
\small
\setlength{\tabcolsep}{2pt}
\begin{tabular}{llccccccc}
\toprule
Model & Method
& Overall$\uparrow$ (\%) & Acc.$\uparrow$ (\%) & Inst.$\uparrow$ (\%) & Cit.$\uparrow$ (\%) & Read.$\uparrow$ (\%)
& Comp.$\uparrow$ (\%) & Dep.$\uparrow$ (\%) \\
\midrule

\rowcolor{gray!20}\multicolumn{9}{c}{\textbf{Closed-Source}} \\
\midrule

Gemini-3-Pro-Preview & In-context & 28.62 & 26.09 & 71.74 & 47.83 & \textbf{25.00} & \textbf{22.22} & \textbf{26.52} \\
\midrule

DeepSeek-V3.2 & In-context & 27.98 & \textbf{28.26} & 80.43 & 50.00 & 21.74 & 16.67 & 26.41 \\
\midrule

Claude-Opus-4.6 & In-context & 23.81 & 23.91 & 78.26 & 50.00 & 17.39 & 11.11 & 22.3 \\
\midrule

\multirow{3}{*}{GPT-5.2} & In-context & 23.61 & 23.91 & \textbf{100.00} & 36.96 & 14.13 & 16.67 & 19.00 \\
 & Mem0 & \textbf{28.65} & 21.74 & \textbf{100.00} & \textbf{76.09} & 20.65 & \textbf{22.22} & 25.30 \\
 & A-MEM & 21.80 & 19.57 & 95.65 & 30.43 & 16.30 & 16.67 & 14.89 \\

\midrule

\rowcolor{gray!20}\multicolumn{9}{c}{\textbf{Open-Source}} \\
\midrule

\multirow{3}{*}{Qwen3-VL-32B-Instruct} & In-context & 14.02 & 13.04 & 80.43 & 19.57 & 3.26 & 16.67 & 10.89 \\
 & Mem0 & 16.74 & 17.39 & 54.35 & 39.13 & 9.78 & 11.11 & 15.00 \\
 & A-MEM & 17.27 & 15.22 & 71.74 & 23.91 & 11.96 & 16.67 & 12.22 \\

\bottomrule
\end{tabular}
\caption{QA results in Type Help. The columns \textit{Acc.}, \textit{Inst.}, \textit{Cit.}, \textit{Read.}, \textit{Comp.}, and \textit{Dep.} stand for Acc Judge score, Inst Judge score, Cit Judge score, Read Judge score, Comprehension score, and Depth score, respectively. $\uparrow$ indicates higher is better. The highest scores are marked in \textbf{bold}.}
\label{tab:qaTYPE}
\end{table*}

\subsubsection{QA Results in No Case Should Remain Unsolved}
In the No Case Should Remain Unsolved scenario, we construct 27 open-ended QA pairs based on the logical linkage of the game. 
The results are reported in Table \ref{tab:qaDUST}.

Overall, closed-source models consistently outperform open-source counterparts across most dimensions, indicating stronger capabilities in maintaining and utilizing long-range evidence under complex investigative settings.  
Among them, Claude-Opus-4.6 achieves the best overall performance, showing a more balanced strength across answer correctness, evidence grounding, and comprehension, while other models exhibit more uneven behaviors across dimensions.
A notable pattern is that most models maintain relatively high instruction-following and citation consistency, yet still struggle with answer accuracy and deep comprehension. 
This suggests that models are often able to retrieve or align with relevant evidence formats, but fail to fully integrate distributed clues into coherent reasoning chains.

\begin{table*}[t]
\centering
\small
\begin{tabular}{lcccccccccc}
\toprule
Model & \makecell{Common \\ Nodes} & \makecell{Human \\ Edges} & \makecell{Model \\ Edges} & TP & FP & FN & Precision$\uparrow$ & Recall$\uparrow$ & F1$\uparrow$ & Jaccard$\uparrow$ \\
\midrule
Gemini-3-Pro-Preview    & 8 & 7 & 4 & 3 & 1 & 4 & \textbf{0.7500} & 0.4286 & 0.5455 & 0.3750 \\
DeepSeek-V3.2           & 8 & 7 & 5 & 3 & 2 & 4 & 0.6000  & 0.4286 & 0.5000    & 0.3333 \\
Claude-Opus-4.6         & 8 & 7 & 11 & 6 & 5 & 1 & 0.5455 & \textbf{0.8571} & \textbf{0.6667} & \textbf{0.5000} \\
GPT-5.2                 & 8 & 7 & 3 & 2 & 1 & 5 & 0.6667 & 0.2857 & 0.4000 & 0.2500 \\
\bottomrule
\end{tabular}
\caption{Example ETD comparisons between human and evaluated models in a selected subsection of Type Help. $\uparrow$ indicates higher is better. The highest scores are marked in \textbf{bold}.}
\label{tab:evalETD}
\end{table*}

\subsubsection{QA Results in Type Help}
In the Type Help scenario, we construct 46 open-ended QA pairs based on the the multi-connection trait within the files, roles and locations.
The results are reported in Table \ref{tab:qaTYPE}.

Compared to previous scenarios, overall performance drops noticeably, indicating that models struggle more when reasoning requires precise aggregation over system-like interactions rather than narrative contexts.
Among all the models, Gemini-3-Pro-Preview achieves the highest Overall score (28.62\%), followed by DeepSeek-V3.2 (27.98\%) and GPT-5.2 with Mem0 (28.65\%). 
A consistent pattern is that most models maintain relatively strong instruction-following and citation behaviors, yet exhibit clear weaknesses in comprehension and depth. 
This indicates that models can often retrieve or align with locally relevant evidence, but fail to construct coherent reasoning chains that integrate multiple interdependent clues. 
In other words, the bottleneck shifts from \textit{finding} information to \textit{structuring} it for multi-step inference.

\subsection{Progress Charts}\label{ProgessCharts}
\subsubsection{Progress Chart in No Case Should Remain Unsolved}
\begin{figure*}[t]
    \centering
    \includegraphics[width=0.95\linewidth]{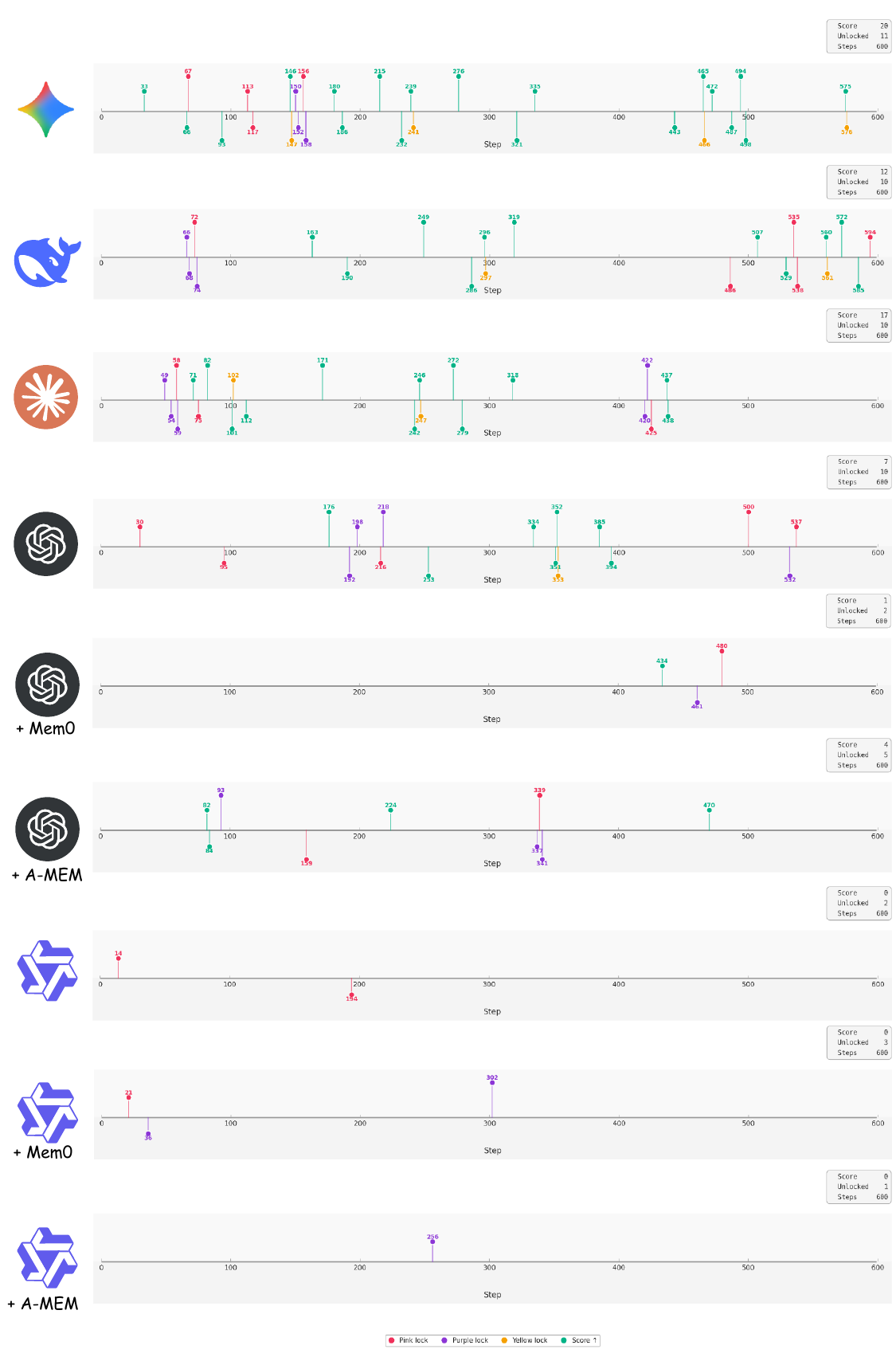}
    \caption{The progress chart of all evaluated models and memory agents in No Case Should Remain Unsolved. The pink, purple, yellow, and green points stand for the model unlocking a pink-lock file, unlocking a purple-lock file, unlocking a yellow-lock file, and sorting 2 memory fragments in correct timeline, respectively.} 
    \label{fig:dustCHART}
\end{figure*}

Figure \ref{fig:dustCHART} illustrates the progress trajectories of different models during the interaction process in \textit{No Case Should Remain Unsolved}, which reflect how models gradually discover memory fragments and how their progress evolves as more interaction steps are taken.
The \textbf{pink lock} assesses the ability to match evidence, the \textbf{purple lock} assesses the ability to deduce and pinpoint specific time points, and the \textbf{yellow lock} assesses the ability to reconstruct the sequence of multiple events. Each model runs for a maximum of 600 steps.


Overall, most models exhibit a gradual and incremental unlocking pattern, where early progress is relatively smooth but later stages become increasingly stagnant. 
This suggests that while initial evidence can be accessed through shallow exploration, advancing further requires successfully connecting previously discovered fragments, which poses a greater challenge.
Many models stop making meaningful progress after reaching a certain point, indicating difficulty in bridging disconnected clues or identifying the next critical evidence node. 
This stagnation reflects a breakdown in long-range reasoning, where models fail to leverage accumulated information to guide subsequent exploration.

\subsubsection{Progress Chart in Type Help}
\begin{figure*}[t]
    \centering
    \includegraphics[width=0.95\linewidth]{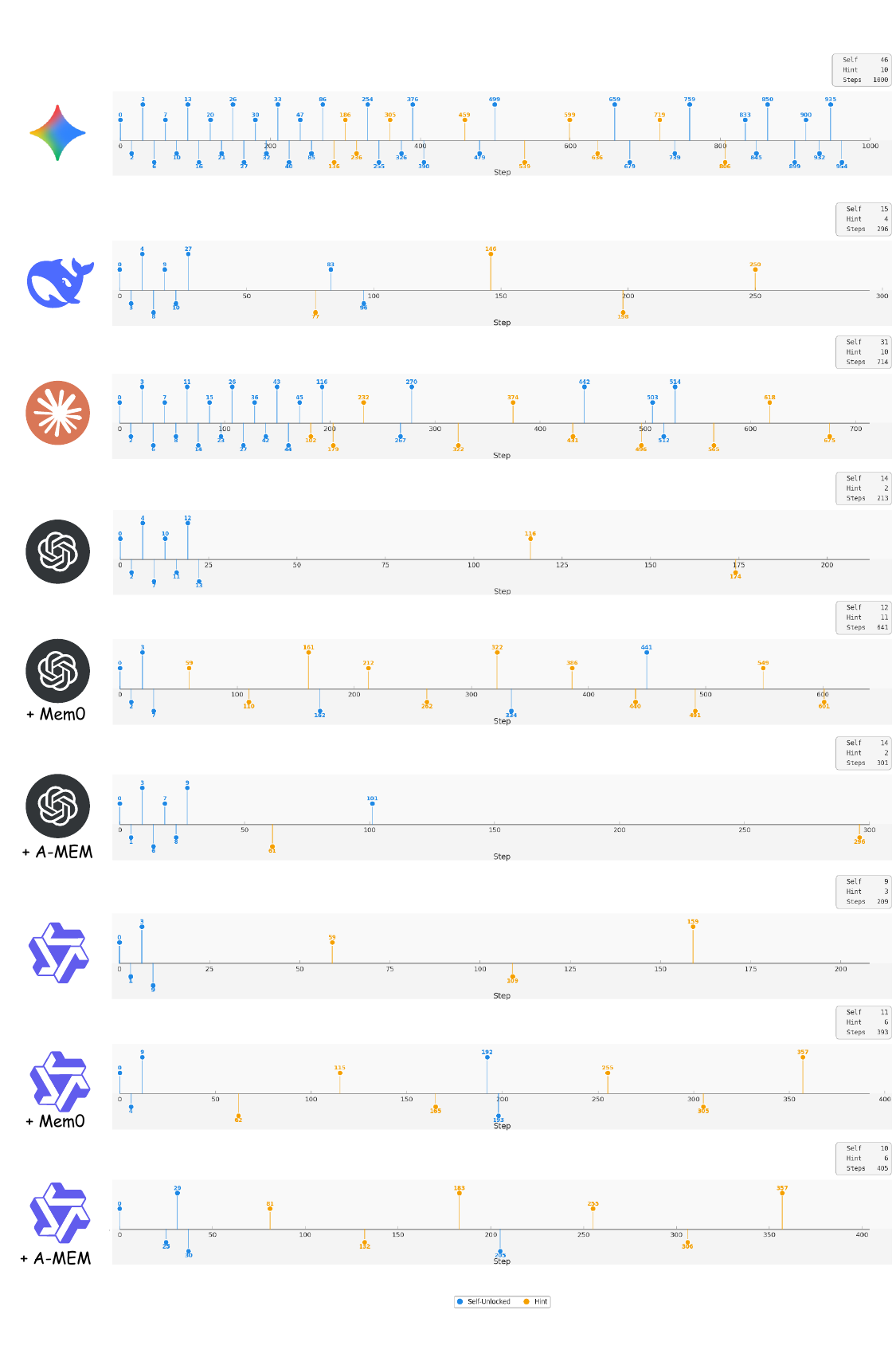}
    \caption{The progress chart of all evaluated models and memory agents in Type Help. The blue and yellow points stand for the model unlocking a file by itself and unlocking a file by receiving a hint, respectively.} 
    \label{fig:typeCHART}
\end{figure*}

Figure \ref{fig:typeCHART} shows the fragment discovery trajectories of different configurations in the \textit{Type Help} scenario. 
Each marker indicates the interaction step at which a specific memory fragment is unlocked during exploration.
If the model hasn’t unlocked a new document for 50 steps, MemGround will provide \textbf{a hint on how to unlock the most recently unlockable document}.
Each model runs for a maximum of 1000 steps.
If the model fails to unlock a new file for 200 consecutive steps, its execution will be terminated.

A prominent pattern is that models frequently enter extended periods of stagnation, during which no new fragments are unlocked despite continued interaction steps. 
These flat segments indicate that the models are unable to identify the correct actions or dependencies required to progress further.
Notably, progress often resumes only after the framework provides explicit hints, which act as external guidance to redirect the exploration process. 
This alternating pattern of stagnation and hint-driven recovery highlights a key limitation: models struggle to infer the underlying structural dependencies between fragments on their own. 
Instead of maintaining a self-directed exploration trajectory, they rely on external signals to overcome local dead-ends.



\subsection{Example ETD Evaluations}\label{DAG}
In this section, we will explain the results from Figure \ref{fig:dag} in detail.
We initially select a subsection of Type Help to evaluate the ETD results of both human and evaluated models, which involves nodes with the IDs ranging from 13 to 20. 
The evaluated models include Claude-Opus-4.6, DeepSeek-V3.2, Gemini-3-Pro-Preview, and GPT-5.2.
Then, we access the model's logs for this game scenario and retrieve information about its past operations on these nodes. 
We also retrieve human ground truth in preparation to making comparisons of both human acts and models' acts.
After obtaining these necessary information, we count the common nodes (the number of common nodes between the human directed graph and the model directed graph), human edges (the number of edges in the human directed graph) and model edges (the number of edges in the model directed graph) in order to calculate several metrics of graphs, as illustrated in Table \ref{tab:evalETD}.

In Table \ref{tab:evalETD}, we first use common nodes, human edges and model edges to count the TP, FP, and FN, which stand for the number of edges common to both graphs, the number of edges present in the model directed graph but not in the human directed graph, and the number of edges present in the human directed graph but not in the model directed graph, respectively.
Then, we apply TP, FP and FN to obtain the scores of Precision, Recall, F1 and Jaccard.
Precision indicates the proportion of edges correctly predicted by the model. Formally:
\begin{equation}
    Precision = \frac{TP}{TP + FP}
\end{equation}
Recall refers to the proportion of edges in the human directed graph that the model correctly identifies. Formally:
\begin{equation}
    Recall = \frac{TP}{TP + FN}
\end{equation}
F1 is the harmonic mean of precision and recall, which can be described as:
\begin{equation}
    F1 = \frac{2\times Precision\times Recall}{Precision + Recall}
\end{equation}
Jaccard represents the ratio of the intersection to the union of the two sets of edges. Formally:
\begin{equation}
    Jaccard = \frac{TP}{TP + FP + FN}
\end{equation}

Table \ref{tab:evalETD} illustrates that even when models identify a largely overlapping set of entities with human annotations, they exhibit substantial variability in constructing the relationships between them. 
Some models tend to produce overly sparse structures, missing critical connections and thus failing to capture the underlying narrative logic, while others generate denser graphs that introduce spurious or weakly supported edges. 
This imbalance between under- and over-connection suggests that the main bottleneck is not entity recognition, but the ability to infer precise relational dependencies from distributed evidence. 
As a result, higher recall in edge construction does not necessarily translate to better structural fidelity, highlighting the challenge of achieving both completeness and correctness in event trajectory modeling.

\section{Case Study}

\subsection{QA Cases}
\subsubsection{Case 1: Fact Transplantation}

\begin{tcolorbox}[title=Case 1]
Scenario: Type Help
\vspace{3pt}

Model: Gemini-3-Pro-Preview
\vspace{3pt}

Question: \textit{What item did the visitor, who was guided to the villa by Harry Thornton, find near the body?}

\vspace{3pt}

Gold Answer: \textit{Chapel key.}
\vspace{3pt}

Model Answer: \textit{The invitation.}
\vspace{3pt}

Key Evidence Source: \textit{T06\_05-DI-1-5}
\vspace{3pt}

Relative Paragraphs:
\vspace{3pt}

"\textbf{John Hobbes} finds a \textbf{key} near the \textbf{corpse} and believes it may not belong to anyone in the house."

"\textbf{Martha Galley} confirms the key is for the \textbf{Chapel}."
"Martha Galley asks John Hobbes to hand the key to the butler \textbf{Harry Thornton} in the Study."

\vspace{3pt}

\tcblower

Model's Reasoning Chain:
\begin{itemize}
    \item Identifies visitor as John Hobbes. \greencheck
    \item Traces visit to Rupert Galley's invitation. \greencheck
    \item Infers "the invitation was found during the search". \redxmark
\end{itemize}

Analysis:
The model transplanted a real fact (Hobbes carried an invitation) into the wrong question context, using a correct clue to fill a gap it never read. The Chapel key evidence was buried in the deeper node T06, which the model never reached; it short-circuited the reasoning chain with "the invitation" instead.
\end{tcolorbox}

\subsubsection{Case 2: Cross-Plot Hallucination}
\begin{tcolorbox}[title=Case 2]
Scenario: No Case Should Remain Unsolved
\vspace{3pt}

Model: GPT-5.2

\end{tcolorbox}

\begin{tcolorbox}[title=]
Question: \textit{After Choi Donggeon's wife passed away, how did the two children (Siwon and Myeong-ho) respectively cope with or adapt to life without their mother?}

\vspace{3pt}

Gold Answer: \textit{Siwon grew dependent on the nanny Shin Huijing and always called her "mom"; Myeong-ho always carried an umbrella outside, honoring a deathbed promise his mother made.}
\vspace{3pt}

Model Answer: \textit{Siwon found comfort in believing the "mom watches over us from heaven" story; Myeong-ho saw through the "kind lie" and fell into moral conflict and self-blame.}
\vspace{3pt}

Key Evidence Source: \textit{D51 "Ms. Shin Hui-jing is", D50 "You mean the umbrella"}
\vspace{3pt}

Relative Paragraphs:
\vspace{3pt}

\textbf{D51:} 
"After my wife suddenly passed away, it was also because of her that the children were able to get through that difficult time smoothly."

"Especially \textbf{Xiyuan} — she was very close to \textbf{Huijing} and always \textbf{called her mom}."
\vspace{3pt}

\textbf{D50: }
"Whether he goes to school or out to play, as long as he leaves the house, he always holds an \textbf{umbrella}."
"He started doing this after my \textbf{wife passed away}."
"\textbf{If one day it rains and you forget to bring an umbrella, I will make the rain stop}, okay?"

\vspace{3pt}

\tcblower

Model's Reasoning Chain:
\begin{itemize}
    \item Reads Myeong-ho's dialogue as internalizing the "kind lie" motif. \greencheck
    \item Reads log as saying "Xiyuan still doesn't know that mom lied". \greencheck
    \item Concludes Siwon copes via belief in heaven, Myeong-ho via moral conflict. \redxmark
    \item Never accesses nanny (D51) or umbrella (D50) nodes. \redxmark
\end{itemize}

Analysis:
The model retrieved and integrated genuine textual content (references to Xiyuan's belief and Myeong-ho's moral doubt), yet applied it to the wrong question frame, producing an answer that is thematically plausible but evidentially displaced.

\end{tcolorbox}

\begin{tcolorbox}[title=]
The model never accessed the nodes containing the nanny (D51) or the umbrella (D50), leaving the correct behavioral evidence entirely outside its context window. 
The result is a complete substitution of the target narrative with an adjacent one, executed with confident causal reasoning.
\end{tcolorbox}

\subsubsection{Case 3: Abstract Character Modeling}

\begin{tcolorbox}[title=Case 3]

Scenario: TRPG
\vspace{3pt}

Model: GPT-5.2
\vspace{3pt}

Question: \textit{How does player5's (Long'er) attitude towards Rumia differ significantly from the other characters, and what does this hint about his level of awareness?}
\vspace{3pt}

Gold Answer: \textit{Long'er responds with calm indifference ("dead fish eyes"), rationally notes that Rumia is not his target, critiques the rule of fairies preying on humans, and directly questions whether Rumia could defeat the Hakurei shrine maiden — suggesting deep understanding of Gensokyo's ecology and power hierarchy.}
\vspace{3pt}

Model Answer: \textit{Long'er approaches Rumia pragmatically — analyzes rules, counters ability, moves on — hinting at a higher awareness of Gensokyo's incident dynamics.}
\vspace{3pt}

Key Evidence Source: \textit{02\_Tracking\_Spring\_Energy\_First\_Encounter\_and\_}

\textit{Probing\_conversation}
\vspace{3pt}

Relative Paragraphs:
\vspace{3pt}

[40] player5: "\# \textbf{Long'er}, with \textbf{dead-fish eyes}, sighs helplessly."

[61] player5: "The \textbf{blonde lady} I'm looking for definitely \textbf{isn't you}."
(Contrast — [62] player4: "There's, there's a \textbf{monster} ahhhhhhhhh")

[75] player5: "\textbf{Fairies preying on humans} is the \textbf{rule} in \textbf{Gensokyo}."

[79] player5: "But I think you probably \textbf{can't beat the Hakurei head}."

\vspace{3pt}

\tcblower

Model's Reasoning Chain:
\begin{itemize}
    \item Frames Long'er as treating Rumia as a solvable "incident obstacle". \greencheck
    
\end{itemize}

\end{tcolorbox}

\begin{tcolorbox}[title=]
\begin{itemize}
    \item Cites teacup echo and causality "stutter" as evidence. (from a different scene, not in D02:40–79) \redxmark
    \item Contrasts with Xueyu's terror, Lina's fear, Ge Qing's refusal. \greencheck
    \item Concludes Long'er has higher genre-knowledge and situational literacy about Gensokyo. \greencheck
    
\end{itemize}

Analysis:
The model arrives at a correct and well-reasoned conclusion while citing evidence that does not correspond to the 

specified source nodes.
The reasoning chain imports plot details from a different scene entirely (the teacup echo and the causality "stutter" ability) neither of which appears in the evidence set D02:40–79.
This suggests the model did not perform targeted retrieval from the cited nodes but instead synthesized a character-level abstraction of Long'er from the broader dialogue context, then projected that abstraction onto the question.
\end{tcolorbox}

\subsection{Evaluation Process Cases}

\subsubsection{Case 4: Hallucinated Recall, Intact Narrative Inference}

\begin{tcolorbox}[title=Case 4]
Scenario: Type Help
\vspace{3pt}

Model: Gemini-3-Pro-Preview
\vspace{3pt}

Phenomenon:
The model successfully unlocked node \textit{04-BI-6-9}, yet its recall field simultaneously cited two nodes, \textit{06-ST-1-11} and \textit{06-ST-1-5-11}, that were neither unlocked nor valid predecessor events in the current narrative. 
These constitute hallucinated references. 
The correct prediction derived not from the recall content, but from the model's understanding of local narrative continuity.
\vspace{3pt}

Key Evidence Source: \textit{T03\_03-LI-1-4-5-6-7-8-9}
\vspace{3pt}

Relative Paragraphs:
\vspace{3pt}
\end{tcolorbox}

\begin{tcolorbox}[title=]

"Person No. 9 demands that \textbf{Edmund Galley} go to the \textbf{Billiard room} to talk \textbf{one-on-one}, emphasizing that he must come."
"It ends by stating that everyone leaves the \textbf{Living room}."
\vspace{8pt}

Analysis:
This case presents a dissociation between two reasoning components. 
The model's explicit memory retrieval mechanism (the recall field) malfunctioned, producing references to nodes that do not exist in the current game state. 
However, its implicit narrative reasoning remained intact: 
the model correctly identified the causal dependency between the demand issued in \textit{03-LI} and its expected execution at the subsequent time node, and further used the failure of \textit{06-BI-6-9} as a contrastive signal to shift the temporal index from \textit{06} to \textit{04}. 
The final prediction is correct despite being grounded in partially fabricated intermediate context. 
This pattern suggests that surface-level memory retrieval and deeper narrative comprehension may operate as partially independent processes, with the latter capable of compensating for failures in the former.
\end{tcolorbox}

\subsubsection{Case 5: Emotional Arc as Temporal Proxy}
\begin{tcolorbox}[title=Case 5]
Scenario: No Case Should Remain Unsolved
\vspace{3pt}

Model: Gemini-3-Pro-Preview
\vspace{3pt}

Phenomenon:
The model achieved its first score (score: 2) by correctly ordering four testimony fragments attributed to Song Minyoung. 
In the absence of explicit timestamps, the ordering was inferred from the directional shift in the character's affective state across fragments.
\vspace{3pt}

Key Evidence Source: (1) \textit{"Hello, you're Siwon's"}, (2) \textit{"Enrollment notice recei"}, (3) \textit{"Calm down first"}, (4) \textit{"You called Siwon's dad"}
\vspace{3pt}

Relative Paragraphs:
\vspace{3pt}

(1) "Hello, you're Siwon's" (Initial contact, calm)

\end{tcolorbox}

\begin{tcolorbox}[title=]
"I thought it was a delivery and it startled me. Next time if you come over for something, could you not ring the \textbf{doorbell}? You can just knock. The child is sleeping."

(2) "Enrollment notice recei" (Cooperating with investigation, neutral)

"Did you receive the \textbf{admission notice}? I heard you didn't go to register."

(3) "Calm down first" (Emotional collapse, child missing)

"That day you came to our home to look for \textbf{Xiyuan}, remember? I even told you not to ring the doorbell... \textbf{Xiyuan is missing!}"

(4) "You called Siwon's dad" (Aftermath, contacting father)

"I've already called him several times, but he still hasn't answered."

\vspace{8pt}

Analysis:
Rather than relying on explicit temporal markers, the model employed the directional trajectory of a character's emotional state as a substitute ordering criterion, a strategy that presupposes sensitivity to affective narrative structure. 
The model's own decision rationale explicitly articulates this logic: "For Gwijja, her shifting attitude from dismissive to worried defines the timeline." 
Beyond emotional arc, the model also appears to have leveraged the recurrence of the "doorbell" detail across fragments (1) and (3) as a cross-segment continuity signal, suggesting that it tracks not only affective state but also surface-level lexical coherence as evidence for narrative ordering. 
This represents a higher-order reading strategy that goes beyond pattern matching toward a form of implicit narrative theory of mind.
\end{tcolorbox}

\end{document}